\documentclass[10pt,letterpaper]{article} 
\usepackage[margin=3.5cm]{geometry}
\usepackage[numbers]{natbib}

\usepackage[noend, ruled, linesnumbered, noline, longend]{algorithm2e}
\usepackage{amsmath,amsthm,amssymb}
\usepackage{siunitx}
\usepackage{tikz}
\usepackage{tikz}
\usepackage{float}
\usepackage{booktabs}
\usepackage{multirow}
\usepackage{subcaption}
\usepackage{calc}
\usetikzlibrary{shapes,arrows,positioning,calc,intersections, decorations.pathreplacing, quotes}
\usepackage{mathtools}
\usepackage{MnSymbol}
\usepackage{todonotes}

\DeclareMathOperator*{\argmin}{arg\,min}
\usepackage[noend, ruled, linesnumbered, noline, longend]{algorithm2e}
\usepackage{easyReview}

\usepackage[automake,acronym,nomain,nopostdot,toc,nowarn]{glossaries-extra}

\makeglossaries

\setabbreviationstyle[jargon]{long-short}
\glssetcategoryattribute{jargon}{glossdescfont}{emph} 

\setabbreviationstyle[acrojargon]{short-nolong}
\glssetcategoryattribute{acrojargon}{glossdescfont}{emph} 

\glsdisablehyper

\newabbreviation[category=jargon]{app}{APP}{\emph{algebraic path problem}}
\newabbreviation[category=jargon]{dag}{DAG}{\emph{directed acyclic graph}}
\newabbreviation[category=jargon]{mosp}{MOSPP}{\emph{multi-objective shortest path problem}}
\newabbreviation[category=jargon]{bdd}{BDD}{\emph{binary decision diagram}}
\newabbreviation[category=jargon]{spc}{SPPC}{\emph{shortest path problem with disjunctive conflicts}}
\newabbreviation[category=jargon]{fpsp}{SPAFPP}{\emph{shortest path avoiding forbidden pairs problem}}
\newabbreviation[category=jargon]{fpspdec}{PAFPP}{\emph{path avoiding forbidden pairs problem}}
\newabbreviation[category=jargon]{fpptfr}{FPPTFR}{\emph{flight planning problem with traffic flow restrictions }}
\newabbreviation{mip}{MIP}{mixed-integer programming}
\newabbreviation{lp}{LP}{linear programming}
\newabbreviation{cp}{CP}{constraint programming}
\newabbreviation{csp}{CSP}{constraint satisfaction problem}
\newabbreviation[category=jargon]{cusp}{CuSP}{\emph{cumulative scheduling problem}}
\newacronym{wrt}{w.r.t.}{with respect to}
\newabbreviation{sat}{SAT}{satisfiability}
\newabbreviation[category=jargon]{cdcl}{CDCL}{\emph{conflict-driven clause learning}}
\newabbreviation{dpll}{DPLL}{Davis-Putnam-Loveland-Logemann}
\newabbreviation{atc}{ATC}{air traffic control}
\newabbreviation[category=jargon]{cnf}{CNF}{\emph{conjunctive normal form}}
\newabbreviation[category=jargon]{dnf}{DNF}{\emph{disjunctive normal form}}
\newacronym{ram}{RAM}{random access memory}
\newabbreviation{dp}{DP}{dynamic programming}
\newabbreviation[category=jargon]{dfa}{DFA}{\emph{deterministic finite automaton}}
\newabbreviation{or}{OR}{operations research}
\newabbreviation{gcd}{gcd}{great-circle distance}
\newabbreviation[category=jargon]{cip}{CIP}{\emph{constraint integer programming}}
\newabbreviation{sp}{SP}{shortest path}
\newabbreviation{spp}{SPP}{shortest path problem}
\newabbreviation[category=jargon]{sppfpaths}{SPPFP}{\emph{shortest path problem with forbidden paths}}
\newabbreviation[category=jargon]{rcspp}{RCSPP}{\emph{resource-constrained shortest path problem}}
\newabbreviation[category=jargon]{tdspp}{TDSPP}{\emph{time-dependent shortest path problem}}
\newabbreviation[category=jargon]{rlcspp}{RLSPP}{\emph{regular language-constrained shortest path problem}}
\newabbreviation[category=jargon]{lbbd}{LBBD}{\emph{logic-based Benders decomposition}}
\newabbreviation[category=jargon]{lcsp}{LCSPP}{\emph{logic-constrained shortest path problem}}
\newabbreviation[category=jargon]{fpp}{FPP}{\emph{flight planning problem}}
\newabbreviation[category=jargon]{tfr}{TFR}{\emph{traffic flow restriction}}
\newacronym{raptor}{RAPTOR}{round-based public transit routing}
\newacronym{np}{NP}{non-deterministic polynomial-time solvable}
\newabbreviation{uk}{UK}{United Kingdom}
\newabbreviation[category=jargon]{up}{UP}{\emph{unit propagation}}
\newabbreviation[category=jargon]{dup}{DUP}{\emph{deep unit propagation}}
\newabbreviation[category=jargon]{sup}{SUP}{\emph{shallow unit propagation}}
\newabbreviation{bb}{B\&B}{branch and bound}
\newabbreviation[category=jargon]{vsids}{VSIDS}{\emph{variable state independent decaying sum}}
\newglossaryentry{lpa}{
  name        = {$\text{LPA}^*$},
  text        = {$\text{LPA}^*$},
  description = {Lifelong Planning $\mathrm{A}^*$},
  sort        = {lpa}
}
\newglossaryentry{lpan}{
    name        = {$\text{LPA}^*$-N},
    text        = {$\text{LPA}^*$-N},
    description = {$\mathrm{LPA}^*$ with Narvaez trick},
    sort        = {lpan}
}
\newabbreviation{lhsy}{LHSY}{Lufthansa Systems Gmbh}
\newabbreviation{toc}{TOC}{top of climb}
\newabbreviation{dfs}{DFS}{depth-first search}
\newabbreviation[category=jargon]{cvds}{CVDS}{\emph{conflict variable decaying sum}}
\newabbreviation[category=jargon]{moms}{MOMS}{\emph{maximum occurrence in minimum size clause}}
\newabbreviation{fifo}{FIFO}{first-in first-out}
\newabbreviation[category=jargon]{ptp}{PTP}{\emph{price-based target pruning}}
\newabbreviation[category=jargon]{fss}{FSS}{\emph{fare-specific speed-up}}
\newabbreviation{ahw}{AHW}{arc-increasing, history-free, and weakly independent}
\newabbreviation{soc}{SoC}{state of charge}
\newabbreviation[category=jargon]{evspr}{EVSP}{\emph{electric vehicle shortest path problem with discretized recharging}}
\newabbreviation[category=jargon]{wcspr}{WCSP-R}{\emph{weight-constrained shortest path problem with replenishment}}
\newacronym{iata}{IATA}{Internation Air Transport Association}
\newabbreviation{od}{OD pair}{origin-destination pair}
\newabbreviation{natots}{NAT-OTS}{North Atlantic Organised Track System}
\newabbreviation{oom}{OOM}{out of memory}

\usepackage{hyperref}
\usepackage[capitalize]{cleveref}

\newtheorem{definition}{Definition}[section]
\newtheorem{proposition}{Proposition}[section]

\newtheorem{remark}{Remark}[section]

\theoremstyle{definition}

\newcommand{\R}{\mathbb{R}}

\newcommand{\G}{G}
\newcommand{\V}{V}
\newcommand{\A}{A}
\newcommand{\weight}{w}
\newcommand{\PS}{P} 
\newcommand{\vrt}{v}

\newcommand{\form}{\Phi}  
\newcommand{\tfrform}{\Phi'}  
\newcommand{\clause}{\alpha} 
\newcommand{\lit}{l}         
\newcommand{\cond}{|}       
\newcommand{\sg}{\rho}      
\newcommand{\trail}{T}      
\newcommand{\confl}{\mathcal{C}} 

\newcommand{\VS}{X}             
\newcommand{\VG}{Y}             
\newcommand{\VF}{Z}             
\newcommand{\vs}{x}             
\newcommand{\vg}{y}
\newcommand{\vf}{z}

\newcommand{\tent}{\mathcal{T}} 

\newcommand{\topsort}{t}           
\newcommand{\outgoing}[1]{\delta^{+}(#1)}  
\newcommand{\ingoing}[1]{\delta^{-}(#1)}   

\newcommand{\viol}{\pi}

\newcommand{\RG}{\G_{\cond\trail}}             
\newcommand{\RD}{\threedgraph_{\cond\trail}}    

\newcommand{\consumption}{\mathtt{consum}}

\newcommand{\cruiseconsumption}[1]{\Psi(#1)}
\newcommand{\duration}{\mathtt{dur}}

\newcommand{\tfrmip}{\texttt{(LCSP)}{}}

\newcommand{\miparcvar}{\lambda}
\newcommand{\mipsatvar}{\mu}

\newcommand{\altitude}[1]{\mathrm{alt}(#1)}
\newcommand{\altdiff}{\Delta}
\newcommand{\speed}{\upsilon}
\newcommand{\climbrate}{\speed_{\mathrm{vert}}}
\newcommand{\projspeed}{\speed_{\mathrm{proj}}}
\newcommand{\twodgraph}{D^{\mathrm{proj}}}
\newcommand{\threedgraph}{D^{3d}}
\newcommand{\threedvrt}{\mathbf{v}}
\newcommand{\threedvrtalt}{\mathbf{u}}

\newcommand{\rhs}{\mathrm{rhs}}

\crefalias{constraint}{equation}
\crefname{constraint}{Constraints}{Constraints}
\Crefname{constraint}{Constraints}{Constraints}
\creflabelformat{constraint}{#2{\upshape(#1)}#3} 

\crefalias{objective}{equation}
\crefname{objective}{Objective}{Objectives}
\Crefname{objective}{Objective}{Objectives}
\creflabelformat{objective}{#2{\upshape(#1)}#3} 

\newcommand{\expnumber}[2]{{#1}\mathrm{e}{#2}}

\title{Logic-Constrained Shortest Paths for Flight Planning}
\date{}
\author{ \href{https://orcid.org/0000-0001-5112-4191}{\includegraphics[scale=0.06]{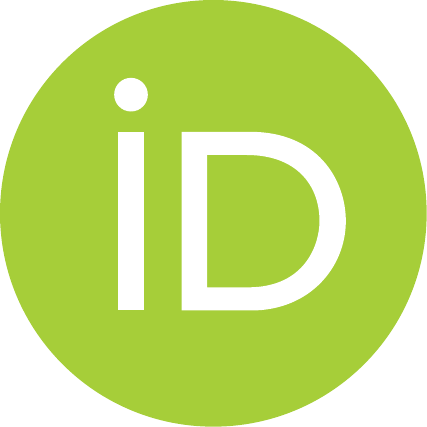}\hspace{1mm}Ricardo Euler}
        \href{https://orcid.org/0000-0002-4197-0893}{\includegraphics[scale=0.06]{orcid_id_icon.pdf}\hspace{1mm}Pedro Maristany de las Casas}\\
        \href{0000-0001-7223-9174}{\includegraphics[scale=0.06]{orcid_id_icon.pdf}\hspace{1mm}Ralf Borndörfer}\\
	Network Optimization \\
	Zuse Institute Berlin\\
	Germany, Berlin, 14195\\
	\texttt{\{euler,maristany,borndoerfer\}@zib.de}\\ }

\usepackage{fancyhdr}
\title{Logic-Constrained Shortest Paths for Flight Planning}
\date{}

\newcommand\fundingfootnote[1]{%
  \let\thefootnote\relax%
  \footnotetext{\textit{#1}}%
  \let\thefootnote\svthefootnote%
}

\begin{document}

\maketitle

\begin{abstract}
The logic-constrained shortest path problem (LCSPP) combines a one-to-one shortest path problem with satisfiability constraints imposed on the routing graph.
This setting arises in flight planning, where air traffic control (ATC) authorities are enforcing a set of traffic flow restrictions (TFRs) on aircraft routes in order to increase safety and throughput. 
We propose a new branch and bound-based algorithm for the LCSPP. 

The resulting algorithm has three main degrees of freedom: the node selection rule, the branching rule
and the conflict.
While node selection and branching rules have been long studied in the MIP and SAT communities, most of them cannot be applied out of the box for the LCSPP. We review the existing literature and develop tailored variants of the most prominent rules.
The conflict, the set of variables to which the branching rule is applied, is unique to the LCSPP. We analyze its theoretical impact on the B\&B algorithm.

In the second part of the paper, we show how to model the flight planning problem with TFRs as an LCSPP
and solve it using the branch and bound algorithm.
We demonstrate the algorithm's efficiency on a dataset consisting of a global flight graph and a set of around 20000 real TFRs obtained from our industry partner Lufthansa Systems GmbH. We make this dataset  publicly available.
Finally, we conduct an empirical in-depth analysis of dynamic shortest path algorithms, node selection rules, branching rules and conflicts.
Carefully choosing an appropriate combination yields an improvement of an order of magnitude compared to an uninformed choice.
\end{abstract}

\fundingfootnote{Funding: This work was supported by the Research Campus MODAL funded by the German Federal Ministry of Education and Research (BMBF) [grant number 05M20ZBM].}

\providecommand{\keywords}[1]{\noindent\textbf{\textit{Index terms---}} #1}
\providecommand{\competinginterest}[1]{\noindent\textbf{\textit{Declaration of interest---}} #1}


\competinginterest{none}

\section{Introduction}\label{lcsp:sec:introduction}

The problem of routing an aircraft from a source airport to a target airport while minimizing the operational cost is called the \gls{fpp} and its multiple variants and constraints have been thoroughly studied in the literature \cite{Blanco2016, Blanco2016a,Blanco2017,Knudsen2017,Knudsen2018, Schienle19, Kuehner2020,Blanco2022}.
The \gls{fpp} is a time-dependent one-to-one shortest path problem defined on a directed graph called an \emph{airway network} with arc cost functions that depend on fuel consumption, weather, and overflight costs.  The two input airports are called an \emph{origin-destination pair} or, in short, an \emph{OD pair}. 
\glsunset{od}

A cornerstone for commercial aircraft routing systems is the handling of so-called \gls{tfr} imposed on the airway network. A \gls{tfr} can, for example, specify that any aircraft that enters Germany using a node on the frontier with Switzerland and heads to the southern part of the UK must leave the European mainland via Bruges. This mandatory rule gives the air traffic controllers some predictability and enables a single controller to survey more aircraft simultaneously. That is why flight planning solvers need to compute optimal routes that adhere to the \gls{tfr} system. Otherwise, the computed routes are not accepted by air traffic controllers.

\begin{figure}[t]
    \centering
    \includegraphics[width=.8\textwidth]{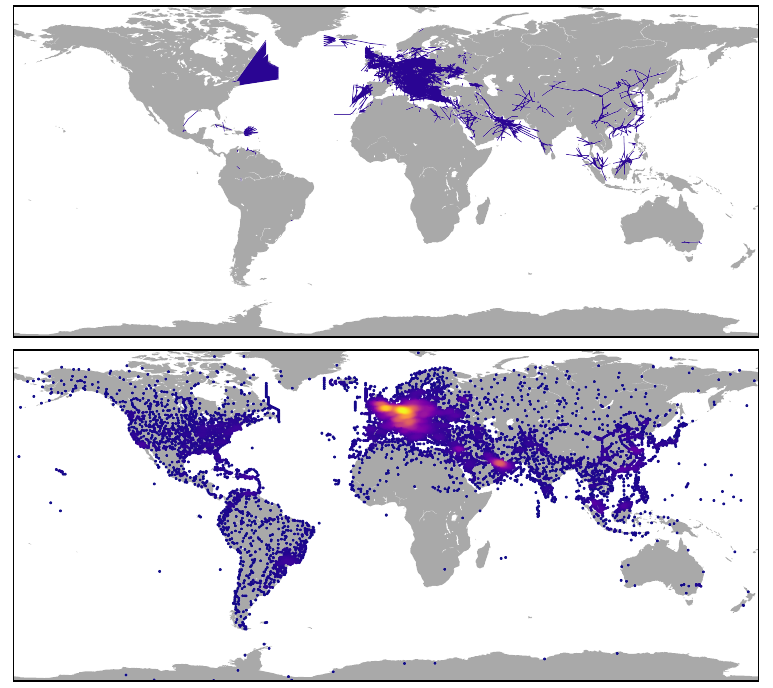}
    \caption{The upper figure depicts all 19300 airway segments that are contained in at least one \gls{tfr}.
    We can see that adherence to the North Atlantic Organized Track System is enforced using \glspl{tfr}.
    The lower figure shows all 11386 waypoints that are contained either directly or via an airway segment in at least one \gls{tfr}. A density distribution of these waypoints in the projective plane was calculated using Gaussian kernel density estimation. It was used to color the waypoints with blue indicating areas with a low density of \glspl{tfr} and red indicating areas with a high density of \glspl{tfr}.}
    \label{lcsp:fig:tfronworld}
\end{figure}

\Glspl{tfr} are stated as propositional formulae that can be formulated in \gls{dnf} where the literals correspond to nodes and arcs in the airway network.
Currently, there are around \num{20000} active \glspl{tfr} (cf. \cref{lcsp:fig:tfronworld}) that are updated every day.

State-of-the-art shortest path algorithms are so called \emph{labeling algorithms}. They build up shortest path trees arc by arc and once they label a node as \emph{settled}, the shortest path up to that node is never changed.
The \gls{tfr} example above exhibits a general algorithmic problem arising in \gls{tfr} handling with labeling algorithms: 
Entering Germany via Switzerland activates a \gls{tfr}, which has consequences only after Bruges, i.e., the resolution of the rule is geographically far away from its activation.
Consider computing a route from Croatia to the UK.
Assume that, at some point during a labeling algorithm, two subpaths $p$ and $q$ meet at a node $v$ in Germany, with $p$ entering the country via Switzerland and $q$ via Austria.
Whenever $v$ is settled in the algorithm, only the cheapest path, say $p$, is stored.
Here, however, this is not possible.
Any path containing  $p$ as a subpath is forced to leave the European mainland via Bruges, whereas $q$ is not restricted in its continuation.
This might result in a more expensive route at the end.
A pure (multi-)labeling approach must hence attach a logical subsystem to each subpath and subpaths only become comparable when their subsystems involve the same \glspl{tfr}.
Due to the many \glspl{tfr} in real-world scenarios, such approaches are forced into storing an exponential number of incomparable paths to be stored until shortly before the target airport is reached, only to keep a single path as an optimal solution.
Consequently, multi-label approaches have been proposed in the literature only for two-dimensional airway networks \citep{Knudsen2017,Kuehner2020} or in a heuristic setting \citep{Knudsen2018}.

\color{black}


In this paper, we model the \glspl{fpptfr} as a \gls{lcsp} and suggest a branch and bound (B\&B) algorithm to solve it.
In every node of the B\&B tree, an \gls{fpp} instance is solved without considering \glspl{tfr}.
The resulting path is then evaluated w.r.t.\ the \glspl{tfr} and the resulting logical infeasibilities are used to branch. 
This black box approach circumvents the above-mentioned memory consumption and running time issues caused by the incomparability of paths.
Across world regions, the distribution of \glspl{tfr} is heterogeneous.
For example, there is a large amount of \glspl{tfr} in Central Europe and the Persian Gulf but very few in Australia or Southern Africa (cf. \cref{lcsp:fig:tfronworld}).
For many \gls{od}, our algorithm thus solves the \gls{lcsp} instance in the root node of the \gls{bb} tree, creating very little overhead compared to a standard \gls{fpp} search.

Since our approach repeatedly recomputes shortest paths, we can make use of one of the several \emph{dynamic shortest path algorithms} in the literature \cite{Ramalingam1996a,Ramalingam1996b,Frigioni2000,Narvaez2000,Koenig2004,Bauer2009,Dandrea2014}. These approaches enable us to warm-start the shortest-path queries in the \gls{bb} tree.
They are meant to avoid the re-exploration of the search space (graph) on, e.g., a route from Australia to Europe, if the only \gls{tfr} violations are close to the destination airport.

\subsection{Related Literature}

The classical shortest path problem and most of its variants are usually addressed by labeling approaches 
\citep[][]{Bellman1958,Dijkstra1959,Floyd1962,Cooke1966,Martins1984,Gandibleux2006,Delling2015,Parmentier2019a,MaristanydelasCasasSedenoNodaBorndoerfer2021}.
When the shortest paths are additionally required to adhere to a set of logical constraints,
applying labeling methods out of the box usually fails, as logical constraints cannot be expressed by manipulation of the graph alone.
For several types of logical constraints, alternative approaches have been studied in the literature \citep{Knudsen2017,Aloul2006,Gabow1976,Pamuk2025,Barrett2000,Nishino2015}.

\citet{Aloul2006} propose a pseudo-Boolean formulation for the shortest path problem, 
which naturally lends itself to the integration of additional logical constraints.
However, even on small graphs and in the absence of logical constraints, they observed poor run times.

More promising are hybrid approaches that enhance labeling techniques with the ability to handle logical constraints.
A precursor of this idea is the \gls{fpsp} \citep{Gabow1976}.
In the \gls{fpsp}, an additional list of forbidden pairs is given as part of the input, and, from each pair, a feasible path may at most contain a single vertex.
The \gls{fpsp} is \gls{np}-hard even on \gls{dag} \citep{Gabow1976}.
Further work on the \gls{fpsp} focuses mostly on complexity results for special cases \citep{Kolman2009, Kovac2013}
or the \gls{fpsp} polytope \citep{Blanco2015}.
\citet{Ferone2021} introduce a general \gls{bb} algorithm for the \gls{fpsp}.
The algorithm works by conducting a standard shortest path search in every tree node.
If the resulting shortest path contains a forbidden pair, two branches are created, in each of which one vertex is removed from the search graph.

The \gls{spc} is a variant of the \gls{fpsp} in which not pairs of vertices are forbidden but pairs of arcs.
Both problems can be polynomially transformed into each other by a transformation of the search graph \citep{Darmann2011}.
\citet{Pamuk2025} give an overview of recent results on the \gls{spc} and \gls{fpsp}.
They also introduce a \gls{bb} algorithm for the one-to-many variant of the problem \citep{Pamuk2025}.

Both the \gls{spc} and the \gls{fpsp} only allow \gls{sat} constraints of the form $\neg a_1 \lor \neg a_2$.
The logical constraints hence form a 2-SAT system, which does not suffice to model general propositional formulae \citep[cf.][]{Aspvall1979,Cook2023}.

\citet{Barrett2000} introduce the \emph{formal-lan\-guage-con\-strained shortest path problem}, in which only paths accepted by an automaton are admissible.
They show the problem to be polynomial-time solvable if the formal language is accepted by a pushdown automaton.
This model finds application in multi-modal passenger routing \citep{Barrett2000,Dibbelt2016}, where the travelers define a preferred modal pattern
which is then translated into an automaton.  
Transforming the input propositional formula of the \gls{lcsp} into an equivalent context-free language is intractable unless $\mathrm{P}=\mathrm{NP}$.
This follows directly from the NP-completeness of \gls{sat}.

\citet{Nishino2015} propose a framework in which logical constraints are formulated as \glspl{bdd}.
Although this approach allows for arbitrary logical constraints, transforming them into a minimum-size \gls{bdd} may still result 
in a diagram of exponential size, relative to the input formula \citep{Bryant1986}. 

A great variety of shortest path problems with additional side constraints have also been studied in the 
\gls{or} literature, usually in the form of the \gls{rcspp} \citep{Desrochers1986}. 
In the \gls{rcspp}, each arc carries a vector of real-valued weights and the weights accumulated along paths are expected to either remain below a given limit 
or to respect time windows at the vertices \citep{Irnich2005}.
Variants of the \gls{rcspp} have among others been tackled using \glsentrylong{dp} \citep{Joksch1966,Dumitrescu2003,Smith2012,Baum2019}, 
backtracking \citep{Lozano2013}, $A^{*}$ search \citep{Barrett2019,Ren2023,Ahmadi2022,Ahmadi2024,Ahmadi2025}, and Lagrangian-based methods \citep{Beasley1989,Borndoerfer2001}.
Surveys are provided by \citet{Irnich2005,Pugliese2013}, and \citet{Festa2015}.

\citet{Irnich2005} introduce a very general model of the \gls{rcspp} that allows for additional \emph{path-structural constraints} that exclude, e.g., $k$-cycles \citep{Irnich2006}. 
They also provide a general \glsentrylong{dp} algorithm that must be adapted by introducing problem-specific dominance rules.
For the \gls{lcsp}, this method is equivalent to the aforementioned multi-labeling approaches and hence suffers from the same weak dominance rules.

The 
\gls{sppfpaths} \citep{Villeneuve2005} maintains a list of forbidden subpaths that cannot appear in any solution, i.e., it only enforces path-structural constraints.
The \gls{lcsp} can be understood as a variant of the \gls{sppfpaths} in which a list of forbidden $s,t$-pairs is implicitly given by the unsatisfying assignments of a propositional formula.
In principle, this viewpoint suggests two possible solution strategies:
First, one could rank paths by their cost using a $k$-shortest path algorithm \citep{Eppstein1998} until a feasible path is found.
While this approach is very flexible, it has already been found to be intractable for the \gls{sppfpaths} \citep{Villeneuve2005}.
Second, one could represent the set of forbidden paths via a \gls{dfa} that is merged with the routing graph, as suggested by \citet{Villeneuve2005}.
Deriving this automaton from a propositional formula is \gls{np}-hard; additionally, the \gls{dnf} may have exponential size in the size of the input formula
\citep{Champarnaud1989}.
Since in the \gls{lcsp} only $s,t$-paths are forbidden, this \gls{dnf} will also not help prune infeasible subpaths early.

The \gls{fpp} has previously been studied thoroughly  in the literature.
A comprehensive overview is given by \citet{Blanco2023}.
\citet{Blanco2016,Blanco2017} extend the \gls{fpp} with overflight costs, which are financial costs airlines must pay for crossing a country's airspace.
Due to the large-scale nature of airway networks, a further focus of research has been on search space reductions \citep{Schienle19}, and 
adapting $\mathrm{A}^*$ search for the complex cost functions of the \gls{fpp} \citep{Blanco2022,Blanco2016a,Knudsen2018}.
The \gls{fpptfr} has already been tackled with multi-labeling approaches.                                                 
These approaches, however, remain either heuristic \citep{Knudsen2018} or restricted to a two-dimensional airway network \citep{Knudsen2017,Kuehner2020}.
More recently, attention has also turned to approaches that no longer rely on an airway network but consider \emph{free flight} \citep{Danecker2022,Borndoerfer2020,Borndoerfer2023,Borndoerfer2023a,Jocas2024,Jensen2017}.
For a thorough review of recent developments, see \citep{Danecker2023}.

\subsection{Contribution}
In this paper, we make three contributions.
First, we formalize the \gls{lcsp} on \glspl{dag} and present a \gls{bb}-based algorithm to solve it.
Second, we adapt several branching strategies from the \gls{mip} and \gls{sat} communities for the \gls{lcsp} and evaluate their performance on a large-scale \gls{fpp}.
Third, we make 
the problem data available to the community. It consists of a realistic world-wide airway network, its corresponding set of \num{18239} \glspl{tfr}, and a simple aircraft model that ensures that realistic optimal routes are computed. 

\begin{remark}
    Airway networks are not acyclic. However, given an \gls{od}, it is a common technique in flight planning to make the routing subgraph acyclic, as arcs pointing in the direction of the departure airport are not needed. 
    This preprocessing is handy when dealing with \gls{lcsp} instances. 
    We need to rule out that paths resolve logical constraints by flying cycles. 
    This is certainly not allowed and would undermine the Air Traffic Controller's motivation to file certain \glspl{tfr}. 
    For example, if flying over a direct connection between two nodes is only allowed after $8$ pm, an aircraft must not cycle around until the so called \texttt{DIRECT} is enabled.
    In the following, we hence consider all directed graphs to be acyclic.
    In \Cref{lcsp:sec:conclusion}, we discuss how our algorithm needs to be adapted to work on graphs with cycles as part of future work.
\end{remark}

\section{An Algorithm for the Logic-Constrained Shortest Path Problem}\label{lcsp:sec:problem_formulation}

We consider a \gls{dag} $\G=(\V,\A)$ and a propositional formula $\form$ over a set of propositional variables $\VS$.
Some variables $\VG\subseteq \VS$ correspond to arcs in $G$ via a bijection $\sg: \A \to \VG$.
We call variables in $\VG$ \emph{graph variables} and variables in $\VF:=\VS\backslash \VG$ \emph{free variables}.

\subsection{Notation for Propositional Logic}


A propositional formula $\form$ is in \gls{cnf} if it is a \emph{conjunction} of \emph{disjunctive clauses}, i.e.
\begin{equation}
    \form = \left( \lit_{11} \lor \dots \lor \lit_{1k_1}  \right) \land \dots \land \left( \lit_{r1} \lor \dots \lor \lit_{rk_r}  \right),
\end{equation}
with $\lit_{ij} \in \bigcup_{\vs\in\VS} \{\vs,\neg \vs\}  \, \forall j\in [k_i] \forall i\in [r]$. 
A propositional formula $\form$ is said to be in \gls{dnf} if and only if it is a \emph{disjunction} of \emph{conjunctive clauses}, i.e.,
\begin{equation}
    \form = \left( \lit_{11} \land \dots \land \lit_{1k_1}  \right) \lor \dots \lor \left( \lit_{r1} \land \dots \land \lit_{rk_r}  \right),
\end{equation}
with $\lit_{ij} \in \bigcup_{\vs\in\VS} \{\vs,\neg \vs\}  \, \forall j\in [k_i] \forall i\in [r]$. 
In the following, we assume $\form$ to be in \gls{cnf}.
This assumption is justified since any propositional formula can be transformed into a \gls{cnf} in linear time using the Tseitin encoding \cite{Tseitin1983}.
The resulting formula contains additional variables representing subformulae, but the size increase is
linear w.r.t.\ the original size of $\form$.
%
The \gls{cnf} formula $\form$ can also be written in set notation as 
\begin{equation}
  \form = \left\{  \left\{ \lit_{ij} : j\in [k_i] \right \}\,:\, i\in [r] \right\}.  
\end{equation}

A set of literals $\trail \subseteq \bigcup_{\vs\in\VS} \{ \vs,\neg \vs \}$ is called 
an \emph{assignment} $\trail$ if it does not contain a contradiction, i.e., for all $\lit \in T$ we have $\neg \lit\not\in T$.
An assignment assigns truth values to the variables $\VS$, i.e., $\vs\in\VS$ is \emph{assigned true} if $\vs \in \trail $ and \emph{assigned false} if $\neg \vs \in \trail$.
If neither $\vs\in \trail$ nor $\neg \vs\in\trail$, the variable $\vs$ is called \emph{unassigned}.
An assignment $\trail$ is \emph{complete} if it assigns all variables, i.e, $|\trail| = |\VS|$.

Using the notation from \citet{Biere2015}, we may \emph{condition} $\form$ on some literal $\lit$ by letting  
\begin{equation}
    \form\cond \lit := \left\{ \alpha \backslash \{\neg \lit\}\ : \ \alpha \in \form, \lit \not\in\alpha \right\}.
\end{equation}
In $\form\cond \lit$, $\lit$ is projected out of $\form$ assuming $\lit$ is true.
This means that all clauses containing $\lit$ are deleted because as disjunctions they are satisfied by $\lit$.
Moreover, $\neg \lit$ is removed from the remaining clauses because $\neg \lit$ will always evaluate to false and hence never satisfy a clause.
All other clauses remain unchanged.
For an assignment $\trail$, we let $\form \cond \trail := \form \cond \lit_1 \cond \dots \cond \lit_k$ for any ordering $(\lit_1,\dots,\lit_k)$ of $\trail$.
The expression is well-defined, since conditioning is order-invariant.

Finally, an assignment $\trail$ \emph{satisfies} $\form$ if $\form \cond \trail = \emptyset$.
Specifically, it satisfies a clause $\clause$ if $\{ \clause \} \cond \trail = \emptyset$.
If there exists no assignment that satisfies $\form$, $\form$ is \emph{unsatisfiable}.
If $\form\cond\trail$ contains the empty clause $\{\}$, there is no assignment $\trail' \supseteq \trail$ that can satisfy $\form$.
In this case, we say that $\trail$ \emph{contradicts} $\form$.      

\subsection{The Logic-Constrained Shortest Path Problem}

The formula $\form$ determines feasibility of paths in $\G$ in the following way:
Every path $p$ \emph{induces} a unique assignment $\trail_p: =  \{\sg(a): a\in p\} \cup \{ \neg \sg(a) : a\not\in p\}$. This assignment assigns all graph variables $Y$ and leaves the free variables $Z$ unassigned.
We say that a path  $p$ and an assignment $\trail$ \emph{agree}
if $p$ induces $\trail \cap \VG$.
The path $p$ \emph{satisfies} $\form$ if there exists a complete assignment $\trail$ that 
agrees with $p$ and satisfies $\form$.

\begin{definition}[The Logic-Constrained Shortest Path Problem] \label{lcsp:def:lcsp}
    \glsreset{lcsp}
    An instance of the \gls{lcsp}, denoted $(\G,\form,\sg,s,t)$, consists of a non-negatively weighted \gls{dag} $\G=(\V,\A,\weight)$ with weights $\weight_a\in\R_{\geq 0}$, $a \in A$, two vertices $s, \, t \in V$, a \gls{cnf} formula $\form$ over propositional variables $\VS = \VG \cupdot \VF$, and a bijection $\sg:\A \to \VG$. A path's cost is the sum of the weights on the path's arcs.
The set of feasible paths from $s$ to $t$ is denoted by $P_{s,t}(\form)$ and contains all $s,t$-paths that satisfy $\form$.
Then, the \gls{lcsp} is to find an $s$-$t$-path $p\in \PS_{s,t}(\form)$ of minimal cost.
\end{definition}

The \gls{lcsp} is \gls{np}-hard. This follows directly from the \gls{np}-hardness of the \gls{fpsp} on \glspl{dag} \citep{Gabow1976}.
Recall that the \gls{fpsp} asks for an $s,t$-path in a graph that does not contain any pair of vertices from a list of pairs 
\begin{equation}
L=\{(a_1,b_1),\dots,(a_k,b_k)\}
\end{equation}
with $a_i,b_i\in \V \, \forall i\in[k]$.
An \gls{fpsp} instance $(\G,L,s,t)$ on a \gls{dag} $\G$ can be transformed in polynomial time into an \gls{lcsp} instance $(\G',\form,\sg,s,t)$ by replacing 
every vertex $v \in V(\G)$  with two vertices $v',v''\in V(\G')$ connected by an arc $v := (v',v'')$ and by replacing every arc $(u,v)\in A(\G)$ with 
$(u'',v')\in A(\G')$.
The list $L$ is converted into the 2-SAT formula
\begin{equation}
\form  = \left\{ \left\{\neg \sg(a_1), \neg \sg(b_1) \right\},\dots, \left\{ \neg \sg(a_k), \neg \sg(b_k) \right\}\right\}.
\end{equation}
The \gls{np}-hardness of the \gls{lcsp} follows directly.

\begin{remark}
    The weights $\weight_a$ in \Cref{lcsp:def:lcsp} are real scalars. When considering the problem in practice, it is most likely that the weights are drawn from (real-valued) functions to capture the evolution of e.g., an aircraft's weight, the trip duration, or the battery charge when traversing paths. Our choice in \Cref{lcsp:def:lcsp} is motivated by the will to benchmark our new algorithm \Cref{lcsp:algo:bb} against the MIP formulation in \Cref{lcsp:eq:mip} (see \Cref{lcsp:sec:experiments}). With scalar weights $\weight_a$ the model stays manageable in size and solvers have a chance to solve small instances.
\end{remark}

A straightforward \gls{mip} formulation of the \gls{lcsp} can be obtained by combining the respective
\gls{mip} formulations of the \gls{spp} and the \gls{sat} problem.
\begin{subequations} \label{lcsp:eq:mip}
    \begin{align}
    \text{\tfrmip} \, & \mathrlap{\min \sum_{(uv)\in\A} \weight_{uv} \miparcvar_{uv}}  \label[objective]{lcsp:mip:objective} \\
    &\text{s.t.}& \sum_{u\in\outgoing{s}} \miparcvar_{us} - \sum_{u \in \ingoing{s}} \miparcvar_{su} &= 1                         \label[constraint]{lcsp:mip:flowa}   \\
    && \sum_{u\in\outgoing{t}} \miparcvar_{ut} - \sum_{u \in \ingoing{t}} \miparcvar_{tu}     &= -1                        \label[constraint]{lcsp:mip:flowb}     \\
    && \sum_{u\in\outgoing{v}} \miparcvar_{uv} - \sum_{u \in \ingoing{v}} \miparcvar_{vu}     &= 0  && \forall \vrt\in\V\backslash\{s,t\}  \label[constraint]{lcsp:mip:flowc}\\
    && \sum_{\substack{\vs\in \VS: \\ \vs\in\alpha}} \mipsatvar_{\vs} + \sum_{ \substack{\vs\in \VS: \\ \neg\vs\in\alpha}} 
    \left(1- \mipsatvar_{\vs}\right)    & \geq 1                              && \forall \clause \in \form  \label[constraint]{lcsp:mip:sat}\\ 
    && \miparcvar_{uv} &= \mipsatvar_{\sg(uv)} && \forall \vg\in\VG \label[constraint]{lcsp:mip:linking} \\
    && \miparcvar_{uv}  &\in [0,1] && \forall (uv)\in\A \\
    && \mipsatvar_{\vs} &\in \{0,1\} && \forall \vs\in \VS.
    \end{align}
\end{subequations} 

\Cref{lcsp:mip:flowa,lcsp:mip:flowb,lcsp:mip:flowc} ensure that, in any feasible solution $(\miparcvar,\mipsatvar)$ to \tfrmip,
the variables $\miparcvar$ induce a shortest $s,t$-path $p$ in $\G$ of weight \mbox{$\weight(p) = \sum_{(uv)\in\A} \weight_{uv}\miparcvar_{uv}$}.
Since $\G$ is acyclic, this holds even though cycle-elimination constraints  are absent from \tfrmip.
\Cref{lcsp:mip:sat} ensure that $\mipsatvar$ satisfies $\form$, while \cref{lcsp:mip:linking} links the 
$\miparcvar$- and $\mipsatvar$-variables, i.e., they ensure that $p$ satisfies $\form$.
Note that the $\miparcvar$-variables are redundant as they are fixed  via \cref{lcsp:mip:linking}.
We only include them for a cleaner presentation.\color{black}

We may now address the \gls{lcsp} by simply solving \tfrmip{} with a commercial \gls{mip} solver. 
However, such an approach fails to leverage the problem's \gls{spp} and \gls{sat} substructures.
Additionally, \tfrmip{} assumes constant arc weights and cannot be easily extended to the \gls{tdspp}.
Many logic-constrained shortest path problems arising in practice, including the \gls{fpp} and \gls{fpptfr}, are however best modeled as a logic-constrained \gls{tdspp}.
We therefore propose a novel \gls{bb} algorithm that is both able to leverage insights from the shortest path and
\gls{sat} communities and that can be easily extended to the case of time-dependent arc costs.

\section{A Branch and Bound Algorithm for the \glsentryshort{lcsp}}\label{tfr:sec:algo}

We derive a \gls{bb} algorithm for the \gls{lcsp} from the following two observations:
First, given an \gls{lcsp} instance $(\G,\form,\sg,s,t)$ and a (partial) assignment $\trail$,
it is possible to construct a subgraph $\RG$ of $G$ in which all $s,t$-paths agree with $\trail$.
This means that if for some $\vg\in\VG$ the literal $\neg \vg$ is in $\trail$ no $s$-$t$-path may contain the arc $\sg^{-1}(\vg)$.
Otherwise, if  $\vg$ is in $\trail$, any $s$-$t$-path in $\RG$ must contain  $\sg^{-1}(\vg)$.
Since there are no cycles in $\G$, 
both cases can be enforced by  deleting a carefully chosen set of arcs from $\G$.
We call this procedure the \emph{enforcement} of $\trail$. It is explained in detail in \cref{lcsp:sec:enforcing}.

Second, $\form$ can be simplified to the equisatisfiable formula $\form\cond\trail$ that contains
no variable assigned by $\trail$.
Combined, this allows us to derive a new \gls{lcsp} instance that assumes all literals in $\trail$ to be assigned
and a corresponding \emph{shortest path relaxation}.

\begin{definition}[]
    \label{lcsp:def:associatedSP}
    Consider an \gls{lcsp} instance as in \cref{lcsp:def:lcsp} and an assignment $\trail{}$ of $\form$. 
    The \emph{graph induced by $\trail$}, denoted by $\RG$, is the subgraph of $G=(V,A)$ obtained by enforcing every arc that is set to true in $T$ and forbidding every arc that is set to false.  We denote by $A(\RG)\subseteq A$ the arc set of $\RG$.
    Then,
    $(\RG,\form\cond\trail,\sg,s,t)$ is a new \gls{lcsp} instance called the \emph{subproblem} induced by $\trail$.
    We call the \emph{one-to-one shortest path} instance $(\RG, s, t)$ the 
    \emph{shortest path relaxation} {of $(\RG,\form\cond\trail,\sg,s,t)$.}
\end{definition}

The algorithm works by repeatedly selecting a subproblem $(\RG,\form\cond\trail,\sg,s,t)$ from a queue of subproblems, generating the subgraph $\RG$, and then solving the shortest path relaxation $(\RG, s, t)$.

\begin{algorithm}
\DontPrintSemicolon
\small
\SetKwInOut{Input}{Input} 
\SetKwInOut{Output}{Output}
\Input{\gls{lcsp} instance $(\G,\form,\sg,s,t)$}
\Output{A cost minimal path $p^*\in P_{s,t}(\form)$}
$p^* \leftarrow \mathrm{NULL}$\tcp*{Incumbent, let $\weight(\mathrm{NULL}) := \infty$}
$Q \leftarrow \{\emptyset\}$\label{lcsp:algo:bb:emptyAssignment}\tcp*{Queue of assignments representing subproblems}
\While{$Q \neq \emptyset$}
{ \label{lcsp:algo:bb:while}
 $\trail \leftarrow Q.\FuncSty{selectNode}()$\tcp*{See \Cref{lcsp:sec:nodeselection}.}\label{lcsp:algo:bb:nodeselection}
 $Q.\FuncSty{deque}(T)$\label{lcsp:algo:bb:deque}\;
$\DataSty{propagated} \leftarrow false$\;
\tcc{
This loop simplifies $\form\cond\trail$ by alternating enforcement and unit propagation until a fixpoint is reached.
}
\While{$\neg \DataSty{propagated}$\label{lcsp:algo:bb:propEnforceLoop}}
{
$\trail \leftarrow \FuncSty{propagate}(\form \cond \trail)$\tcp*{See \Cref{lcsp:sec:propagation}.} \label{lcsp:algo:bb:propagation}
\If(\tcp*[f]{Contradiction found in $\form\cond\trail$}){$\{\} \in \form \cond \trail$}
{
\KwSty{goto} \Cref{lcsp:algo:bb:while}\;
}\label{lcsp:algo:bb:satinfeas}

$\RG \leftarrow \FuncSty{enforce}(G, T)$\tcp*{See \Cref{lcsp:sec:enforcing}.} \label{lcsp:algo:bb:enforce}

\If(\tcp*[f]{Contradiction from from enforcing $\trail$ in $\RG$}){ $\{a \in A\backslash A(\RG) : \sg(a) \in \trail \} \neq \emptyset $}
{\label{lcsp:algo:bb:enforceinfeas}
\KwSty{goto} \Cref{lcsp:algo:bb:while}\;
}

$\mathcal{L} \leftarrow \{a \in A\backslash A(\RG) : \neg \sg(a) \not \in \trail \}$\tcp*{Implications found from enforcing $\trail$ in $\RG$}\label{lcsp:algo:bb:enforceprop1}
$\trail \leftarrow \trail \cup \{\neg\sg(a) : a\in \mathcal{L} \}$\;\label{lcsp:algo:bb:enforceprop2}
\If{$ \mathcal{L} = \emptyset$}
{
$\DataSty{propagated} \leftarrow true$\;
}

}\label{lcsp:algo:bb:propEnforceLoopEnd}

$p \leftarrow \FuncSty{shortestPath}(\RG,s,t)$\; \label{lcsp:algo:bb:sp}
\If{$p \neq  \mathrm{NULL}$ {\upshape and} $\weight(p) < \weight(p^*)$\label{lcsp:algo:bb:pathnull}}
{
$\tent \leftarrow \trail \cup \tent_p $\; \label{lcsp:algo:bb:tent}

\If{SAT($\form \cond \tent$)\label{lcsp:algo:bb:satresolution}} 
{
$p^* \leftarrow p$\;\label{lcsp:algo:bb:updateincumbent}
} 
\Else{
$\vs \leftarrow \FuncSty{chooseVariable}(X\backslash T)$\tcp*{See \Cref{lcsp:sec:mipbranching,lcsp:sec:satbranching}.} \label{lcsp:algo:bb:branching}
$Q.\FuncSty{append}(\trail\cup\{\vs\})$\label{lcsp:algo:bb:newNode1}\;
$Q.\FuncSty{append}(\trail\cup \{\neg \vs \})$\label{lcsp:algo:bb:newNode2}\;
}
}
}
\Return $p^*$\label{lcsp:algo:bb:return}\;
\caption[Branch and bound algorithm for the \glsentryshort{lcsp}]{\Glsentrylong{bb} algorithm for the \glsentryshort{lcsp}.\label{lcsp:algo:bb}}
\end{algorithm}

Clearly, an optimal solution $p$ to $(\RG, s, t)$ may not satisfy $\form \cond \trail$.
When this happens, our algorithm chooses an unassigned variable $\vs\in\VS\backslash\trail$  
and creates two new subproblems $(\G_{\cond \trail\cup\{\vs\}},\form\cond\trail \cup \{\vs\},\sg,s,t)$ and \mbox{$(\G_{\cond \trail\cup\{ \neg \vs\}},\form\cond\trail\cup\{\neg \vs\},\sg,s,t)$}.

Both the variable assignment in the branching step and the enforcement of $\trail$  might cause logical implications in $\form\cond\trail$.
These implications further simplify $\form\cond\trail$ and may even lead to contradictions.
Deriving these implications is called \emph{propagation} and is discussed in detail in \cref{lcsp:sec:propagation}.
Before solving the shortest path relaxation, the algorithm alternates between a propagation and an enforcement step until no further progress can be made.

In the following, we give a detailed description of the complete algorithm.
The pseudocode can be found in \cref{lcsp:algo:bb}.
The subroutines are explained in \cref{lcsp:sec:propagation,lcsp:sec:conflict,lcsp:sec:enforcing}.



\paragraph{Initialization}
The algorithm maintains an incumbent path $p^* \in P_{s,t}(\form)$ 
and, implicitly, a \gls{bb} tree.
Each node in the tree corresponds to an assignment $\trail$ of $\form$.
A queue $Q$ stores the nodes that have not yet been processed.
The root node of the \gls{bb} tree corresponds to the empty assignment, which we denote by $\emptyset$. 
It is pushed to $Q$ in \Cref{lcsp:algo:bb:emptyAssignment}.

\paragraph{Main Loop}
In \cref{lcsp:algo:bb:nodeselection}, an unprocessed assignment $\trail$ is selected from $Q$ using a node selection rule (\cref{lcsp:sec:nodeselection}) and dequeued in \cref{lcsp:algo:bb:deque}.
In \cref{lcsp:algo:bb:propagation}, a propagation heuristic on $\form\cond\trail$ assigns additional literals in $\trail$, thereby simplifying $\form \cond \trail{}$ (see \cref{lcsp:sec:propagation}).
If $\form \cond \trail$  then contains the empty clause $\{\}$, $\form \cond \trail$ is unsatisfiable, certifying infeasibility of the subproblem  $(\RG,\form\cond\trail,\sg,s,t)$.

If no logical infeasibility is detected, we build the shortest path instance $(\RG, s, t)$ associated to $T$ in \cref{lcsp:algo:bb:enforce}.
This is done by deleting arcs from $\G$ that conflict with the literals in $\trail$. The procedure is explained in detail in \cref{lcsp:sec:enforcing}. 
The remaining $s,t$-paths in $\RG$ are precisely those agreeing with $\trail$ (cf. \cref{lcsp:prop:feasiblePaths}).

For any deleted arc $a\in \A \backslash \A(\RG)$, $\sg(a)$ must be assigned false in $\trail$.
If any such arc exists that is assigned true in $\trail$, we have found a contradiction in $\trail$ and the current subproblem is infeasible.
This is checked in \cref{lcsp:algo:bb:enforceinfeas}.
Then, any unassigned variable $\sg(a)$ with $a\in \A \backslash \A(\RG)$  is assigned false in $\trail$ in \cref{lcsp:algo:bb:enforceprop1,lcsp:algo:bb:enforceprop2}.
This may trigger new propagations such that we return to \cref{lcsp:algo:bb:propEnforceLoop}.
This process is repeated until either infeasibility is detected or no new propagations can be made.

Then, the shortest path instance $(\RG, s, t)$ is solved in \cref{lcsp:algo:bb:sp}.
Let $p$ be the solution obtained for $(\RG, s, t)$.
If $p = \mathrm{NULL}$, $s$ and $t$ are disconnected in $\RG$ implying that
there exists no path that agrees with $\trail$.
Again, we find $(\RG,\form\cond\trail,\sg,s,t)$ to be infeasible.
Otherwise, if $\weight(p) < \weight(p^*)$, we compute the union $\tent$ of $\trail$ and the assignment $\trail_p$ induced by $p$ in \cref{lcsp:algo:bb:tent}.  
By construction of $\RG$, $\tent$ contains no contradiction and is hence an assignment.
In \cref{lcsp:algo:bb:satresolution}, we check whether $p\in P_{s,t}(\form\cond\tent)$ .
Recall that $\tent$ contains $\trail_p$ and since $\trail_p$ is an assignment induced by a path, it assigns all graph variables.  %
Hence, $\form\cond\tent$ contains only free variables, and, to determine whether $p\in P_{s,t}(\form\cond\tent)$,
it suffices to solve a pure SAT problem over the formula $\form \cond \tent$.
If $\form \cond \tent$ is satisfiable, $p$ satisfies $\form$, and it becomes the new incumbent $p^*$ (\cref{lcsp:algo:bb:updateincumbent}).
If $\form \cond \tent $ is unsatisfiable, we select any variable $\vg$ not yet in $\trail$ 
and add two new assignments $\trail\cup\{\vg \}$ and $\trail\cup\{ \neg \vg \}$ to the queue (\cref{lcsp:algo:bb:newNode1,lcsp:algo:bb:newNode2}).

\paragraph{Termination}
The algorithm terminates when the queue $Q$ is found to be empty at the beginning of an iteration of the main loop. When this happens, the incumbent path $p^*$ is returned (\cref{lcsp:algo:bb:return}).

\begin{proposition}\label{lcsp:prop:correct}
  \Cref{lcsp:algo:bb} solves the \gls{lcsp}.
 \end{proposition}

\begin{proof}
If the check in \cref{lcsp:algo:bb:pathnull} never fails,  \cref{lcsp:algo:bb} will enumerate all complete assignments $\trail$ that satisfy $\form$ and, for each $\trail$, compute a cost minimal $s,t$-path in $\RG$.
If the check in \cref{lcsp:algo:bb:pathnull} fails, either $\RG$ is disconnected 
or any $s,t$-path in $\RG$ has higher weight than $p^*$. Both 
hold for any  $\G_{\cond\trail'}$ with $\trail'\supset \trail$ as well. Hence, $\trail$ need no longer be considered.
\end{proof} 

\subsection{Propagation}\label{lcsp:sec:propagation}
The goal in \cref{lcsp:algo:bb:propagation} is to strengthen the current subproblem by assigning additional variables in $\trail$ and simplifying $\form\cond\trail$.
To do so, we employ \emph{unit propagation} which is the core propagation technique at the heart of \gls{dpll} \citep{DPLL1962} and \gls{cdcl} \gls{sat} solvers \citep{Silva1996,Moskewicz2001,Ryan2004}.
The technique searches for a \emph{unit clause} in $\form\cond\trail$, i.e, a clause containing only one literal $\lit$.
As adding $\neg \lit$ to $\trail$ would generate the empty clause, we can then
replace $\form\cond\trail$ by $\form \cond \trail\cup\{\lit\}$.
By doing so, clauses containing $\lit$ are fulfilled, and, most importantly, clauses containing $\neg \lit$ decrease in size and may become unit clauses.
This procedure is repeated until no more unit clauses are found.

Unit propagation is \emph{sound}, i.e., it generates new valid clauses but not \emph{refutation-complete}, i.e., it might be unable to produce the empty clause even if $\form \cond \trail$ is unsatisfiable \citep{Handbook2021}.
Depending on the problem structure, it may be worthwhile to employ a refutation-complete resolution method, e.g, \emph{linear resolution} \citep{Loveland1970}.

On free variables in $\VF$, we may also perform \emph{pure literal elimination} \citep{DPLL1962}, i.e., if a literal $\lit$ appears in $\form\cond\trail$ but not its negation $\neg\lit$, we can set $\trail \leftarrow \trail\cup\{\lit\}$.
This technique does not extend to graph variables, as assigning them has an effect on the routing graph $\RG$. 

\subsection{Enforcing Literals and Shortest Path Search}\label{lcsp:sec:enforcing}

In \cref{lcsp:algo:bb:enforce} of \cref{lcsp:algo:bb}, $\FuncSty{enforce}(\G,\trail)$ builds a subgraph $\RG$ of $G$ in which
all $s,t$-paths agree with $\trail$.
This is ensured as follows:
For literals $\neg \vg$, we delete $\sg^{-1}(y)$ from the arc set $\A$.
For literals $\vg$, we enforce $\sg^{-1}(\vg)$ to be contained in every $s,t$-path by deleting a set of alternative arcs.
As $\G$ is acyclic, we compute a topological sorting $\topsort: V\mapsto \mathbb{N}$ of $\V$ in linear time w.r.t.\ $G$'s size \citep{Kahn1962}.
Then, to enforce $\sg^{-1}(\vg) = (u,v)$, we delete all arcs $\outgoing{u}\backslash\{ (u,v) \}$
as well as all arcs $(i,j)$ with $\topsort(i) < \topsort(u)$ and $\topsort(j) > \topsort(u)$. 
For an example, see \cref{lcsp:figure:enforce}.

\begin{proposition}\label{lcsp:prop:enforce}
\label{lcsp:prop:feasiblePaths}
The $s,t$-paths in $\RG$ are exactly those agreeing with $\trail$.    
\end{proposition}
\begin{proof}
For each $\neg \vg \in \trail$, $\sg^{-1}(\vg)$ is deleted in $\RG$.
For each $\vg\in\trail$, $\sg^{-1}(\vg)$ is a bridge separating $s$ and $t$ and thus part of any $s,t$-path in $\RG$.
Hence, all $s,t$-paths in $\G$ agree with $\trail$.
By the topological sorting, no $s,t$-path agreeing with $\trail$ may use one of the deleted edges
and hence all $s,t$-path agreeing with $\trail$ must be in $\RG$.
\end{proof}

In \cref{lcsp:algo:bb:sp}, $\FuncSty{shortestPath}(\RG,s,t)$  
solves the shortest path relaxation $(\RG,s,t)$ by computing 
a shortest $s,t$-path $p$ in $\RG$. 
Here, any shortest path algorithm may be used. 
Note that, to speed up the algorithm, any (sub)path processed in the shortest path algorithm that exceeds the costs of the incumbent $p^*$ can be neglected.

The shortest path search can be performed using a dynamic shortest path algorithm in the following way:
The shortest path search manages a single shortest path tree and keeps a reference to the last assignment $\trail'$ and graph
$\G_{\cond \trail'}$ on which a shortest $s,t$-path was calculated.
When a shortest path search on $\RG$ is triggered for a new assignment $\trail$, all differences between $\RG$ and $\G_{\cond\trail'}$ are processed in the  initialization phase of the algorithm. 
Then, the main phase is started to obtain a shortest path tree for $\RG$.
Since $\trail$ need not be a predecessor of $\trail'$ in the \gls{bb} tree,
the new graph $\G_{\cond\trail}$ may contain \emph{inserted} and \emph{deleted} arcs with respect to $\G_{\cond\trail'}$.
There are several options available in the literature \citep{Ramalingam1996a,Ramalingam1996b,Frigioni2000,Narvaez2000,Koenig2004} that can deal with arc insertions and deletions. The best choice, however, depends heavily on the graph structure \citep{Bauer2009,Dandrea2014} such that no general recommendation can be made here.

Finally, for some assignment $\trail$, the shortest path computed in \cref{lcsp:algo:bb:sp} may be identical to
the path $p_{\mathrm{parent}(\trail)}$ computed in its parent node, denoted by $\mathrm{parent}(\trail)$.
Before running $\FuncSty{shortestPath}(\RG)$, we hence check whether
\begin{equation}\label{lcsp:algo:bb:checkparent}
    p_{\mathrm{parent}(\trail)} \subseteq \A\left(\G_{\cond \trail}\right).
\end{equation}
If yes, $p_{\mathrm{parent}(\trail)}$ is optimal for $\mathrm{parent}(\trail)$ and feasible for $\trail$.
It is hence also optimal for $\trail$, and the shortest path search is skipped.

\begin{figure}[t]
\centering
\begin{tikzpicture}[scale=.65,
        font={\fontsize{8pt}{8}\selectfont},
        arc/.style={-latex, thick},
        delarc/.style={-latex, dashed, thick, color=gray!90},
        node/.style={draw,circle}
        ]
    
    \node[node] (v1) at (0,0) {$s$};
    \node[node] (v2) at (0,4) {$v_3$};
    \node[node] (v3) at (2,2) {$v_2$};
    \node[node] (v4) at (4,0) {$v_5$};
    \node[node] (v5) at (4,4) {$v_4$};
    \node[node] (v6) at (6,2) {$t$};


    \node[node] (sort1) at (8, 2)  {$s$};
    \node[node] (sort2) at (10, 2) {$v_3$};
    \node[node] (sort3) at (12, 2) {$v_2$};
    \node[node] (sort4) at (14, 2) {$v_5$};
    \node[node] (sort5) at (16, 2) {$v_4$};
    \node[node] (sort6) at (18, 2) {$t$};

    \draw[arc] (v1) -- (v2);
    \draw[arc] (v1) -- (v3);
    \draw[delarc] (v1) -- (v4);
    \draw[delarc] (v2) -- (v5);
    \draw[arc] (v2) -- (v3);
    \draw[delarc] (v3) -- (v4);
    \draw[arc] (v3) -- (v5);
    \draw[arc] (v4) -- (v5);
    \draw[delarc] (v3) -- (v6);
    \draw[arc] (v4) -- (v6);
    \draw[arc] (v5) -- (v6);

    \draw[arc] (sort1) edge[] (sort2);
    \draw[arc] (sort1) edge[bend left=30] (sort3);
    \draw[delarc] (sort1) edge[bend right=30] (sort4);
    \draw[delarc] (sort2) edge[bend left=30] (sort5);
    \draw[arc] (sort2) edge[] (sort3);
    \draw[delarc] (sort3) edge[] (sort4);
    \draw[arc] (sort3) edge[bend right=30] (sort5);
    \draw[arc] (sort4) edge[] (sort5);
    \draw[delarc] (sort3) edge[bend right=45] (sort6);
    \draw[arc] (sort4) edge[bend left=30] (sort6);
    \draw[arc] (sort5) edge[] (sort6);

\end{tikzpicture}
\caption[Topological sorting of a directed acyclic graph]{
    Left: \Gls{dag} $D$ in which the arc $(v_2,v_4)$ is enforced. Right:
    Topological sorting of $D$. }
\label{lcsp:figure:enforce}
\end{figure}

\subsection{Validation and Conflict Generation}\label{lcsp:sec:conflict}

The shortest path $p$ computed in \cref{lcsp:algo:bb:sp} induces an assignment $\tent_p$.
By construction of $\RG$, $p$ agrees with $\trail$. 
Hence, it agrees with $\tent := \trail \cup \tent_p$ (\cref{lcsp:algo:bb:tent}).
Since $\tent_p$ assigns all graph variables, $\tent$ does as well, and the formula $\form \cond \tent$ contains only free variables.
If $\form \cond \tent$ is satisfiable, we have hence found a shortest path in $P_{st}(\form \cond \tent) \subseteq P_{st}(\form\cond\trail)
\subseteq P_{st}(\form)$ and can update the incumbent solution $p^*$ in \cref{lcsp:algo:bb:updateincumbent}. 

If $\form \cond \tent$ is unsatisfiable, we need to choose an unassigned variable to branch on.
\cref{lcsp:algo:bb} is correct for any choice from $\VS\backslash \trail$.
It is, however, preferable to identify a subset of variables $\confl \subset \VS\backslash\trail$ that we suspect to be in some way responsible for the unsatisfiability. We call such a set a \emph{conflict}.

The conflict plays a similar role to the set of fractional variables in \gls{mip} solving. 
It is, however, not necessarily unique.
It is also not to be confused with the learned conflicts in \gls{cdcl} solvers, which are formed by variables from $\trail$.

A conflict can be obtained, for example, as the set of all variables occurring in an \emph{unsatisfiable core} of $\form \cond \trail$, that is, a subset of clauses 
that remains unsatisfiable. 
Computing unsatisfiable cores is a standard feature of incremental \gls{sat} solvers like MiniSAT \citep{Een2003,MarquesSilva2021}.
As only graph variables lead to new enforcements in the graph, it might be worthwhile to only consider conflicts consisting of graph variables.

\begin{proposition}\label{lcsp:prop:disjointbranching}
Branching exclusively on graph variables and performing the check in \cref{lcsp:algo:bb:checkparent} guarantees that 
\cref{lcsp:algo:bb} computes no shortest $s,t$-path more than once.
\end{proposition}
\begin{proof}

Let $p$ be a shortest path in both $\G_{\cond\trail_1}$ and $\G_{\cond\trail_2}$ for two assignments $\trail_1,\trail_2$ that were derived from the empty assignment by branching exclusively on graph variables. 
Let $\trail = \trail_1 \cap \trail_2$ be the lowest common ancestor of $\trail_1$ and $\trail_2$ in the \gls{bb} tree.
The path $p$ must be a feasible $s,t$-path in $\G_{\cond\trail}$.

After branching on any graph variable $\vg\in\VG\backslash\trail$, $p$ will be infeasible in at least one of the child nodes $\trail\cup\{\vg\}$ and $ \trail \cup \{\neg \vg\}$.
W.l.o.g. assume this is $\trail\cup\{\vg\}$. 
Since $\G_{\cond\trail'} \subseteq \G_{\cond\trail\cup\{\vg\}}$ for all assignments $\trail'\supseteq \trail\cup\{\vg\}$,
this also holds for any child node of $\trail\cup\{\vg\}$.
This means that all assignments for which $p$ is an optimal shortest path must lie on an oriented path in the \gls{bb} tree.
Therefore, checking \cref{lcsp:algo:bb:checkparent} suffices to avoid recomputations of $p$.  
\end{proof}

 \Cref{lcsp:figure:free_bad} shows that branching on graph variables is a necessary condition in \cref{lcsp:prop:disjointbranching}.

\begin{figure}[ht]
\centering
\begin{tikzpicture}[scale=1,
        font={\fontsize{9pt}{9}\selectfont},
        arc/.style={-latex, thick},
        delarc/.style={-latex, dashed, thick, color=gray!90},
        node/.style={draw,circle}
        ]

      \node[node] (v1) at (0,0) {$s$};
      \node[node] (v2) at (4,0) {$v$};
      \node[node] (v3) at (8,0) {$t$};
      \draw[arc] (v1) edge[] node[above] {$w_{sv}=1$} (v2);
      \draw[arc] (v2) edge[] node[above] {$w_{vt}=1 $}(v3);
      \draw[arc] (v1) edge[bend right=25]node[below] {$w_{st}=5 $} (v3);
\end{tikzpicture}

\caption[Necessity of branching on graph variables]{
Consider the \gls{lcsp} instance $(\G,\form,\sg,s,t)$ given by the above
graph $\G$, 
the \gls{cnf} formula $\form = (\vg\lor\vf) \land (\vg \lor \neg \vf)$
over the variable set $\VS := \VG\cup \VF$ with 
$\VG := \{ o,x,y \}$ and $\VF := \{ z \}$ and the bijection $\sg: \{(sv),(vt),(vt)\} \to \VG$
with $\sg(sv) = o$, $\sg(vt) = x$ and $\sg(st) = y$.
%
The shortest path in $G_{\cond\emptyset}$ is $(s,v,t)$ with a weight of $2$.
However, $p$ induces the assignment $\{\neg \vg \}$ which conflicts with $\form$.
Branching on $\vf$ results in two child assignments $\{\vf\}$ and $\{\neg\vf\}$
and also $\form\cond \{\vf\} = \vg$ and $\form\cond \{\neg\vf\} = \vg$.
Hence, both yield the shortest path $(s,t)$.
}\label{lcsp:figure:free_bad}
\end{figure}

Free variables can be eliminated from a formula $\form$ by \emph{existential quantification} \citep{Handbook2021}:  the formulae $\exists\vf\form := (\form \cond \vf) \lor (\form \cond \neg \vf)$ and $\form$ are equisatisfiable, i.e.,
$\form$ is satisfiable if and only if  $\exists\vf\form$ is satisfiable.
The variable $\vf$ does not appear in $\exists\vf\form$. After obtaining an
assignment $\trail'$  that satisfies $\exists\vf\form$, the value for $\vf$ can be found by checking whether $\trail'$ satisfies $(\form \cond \vf)$ or $(\form \cond \neg \vf)$. If $\trail'$ satisfies $(\form \cond \vf)$,  $\trail:=\trail'\cup\{\vf\}$
satisfies $\form$. Otherwise, $\trail:=\trail'\cup\{\neg \vf\}$ satisfies $\form$.
We can hence obtain a conflict in terms of graph variables by considering  $\exists \left(\VF\backslash\trail\right) \form \cond \tent$ instead of $\form \cond \tent$. 
However, $\exists \left(\VF\backslash\trail\right) \form \cond \tent$ is in general of exponential size w.r.t. the size of $\form\cond\tent$.
We call conflicts $\confl \subset \VG$ \emph{graph conflicts};  all other conflicts are called
\emph{regular conflicts}.

\section{Node Selection and Branching Rules}\label{lcsp:sec:nodeandbranch}
Two important degrees of freedom in \cref{lcsp:algo:bb} are the assignment selected for processing in \cref{lcsp:algo:bb:nodeselection}
and the branching decision in \cref{lcsp:algo:bb:branching}.
It is well known in the \gls{sat} and \gls{mip} communities that selecting the right variable to branch on has a significant impact on the size of the search tree \citep{Achterberg2007, Biere2015}.
Variable choices for branching are referred to as \emph{branching rules} \citep{Achterberg2007} in the \gls{mip} community and as \emph{variable selection heuristics} \citep{Handbook2021} in the \gls{sat} community.
In the following, we recap various node selection and branching rules from both communities and adapt them for the \gls{lcsp}.
In \cref{lcsp:sec:computation}, we evaluate their performance.

We use the following notational convention:
For an assignment $\trail$ that has already been processing in \cref{lcsp:algo:bb}, we denote the path computed in \cref{lcsp:algo:bb:sp} by 
$p_{\trail}$ and by $\viol_\trail$ the number of clauses in $\form$ violated by the assignment $\tent$ computed in \cref{lcsp:algo:bb:tent}.
For a queued, but unprocessed assignment $\trail\in Q$, we let $p_{\trail} := p_{\mathrm{parent}(\trail)}$ and
$\viol_{\trail} :=\viol_{\mathrm{parent}(\trail)}$.
At any point during the execution of \cref{lcsp:algo:bb}, a
\emph{primal bound} is given by $b_p := \weight(p^*)$ and a \emph{dual bound} by
\begin{equation}
b_d := \min\left(
    \weight\left(p^*\right),
    \min_{\trail \in Q}\left(\weight\left(p_{\trail}\right)\right)\right).
\end{equation}

\subsection{Node Selection}\label{lcsp:sec:nodeselection}

{\Gls{dfs}} \citep{Dakin1965}, selects a child of the current node. If there is none, it backtracks until a child is found.
\Gls{dfs} aims to quickly improve the primal bound as feasible solutions usually appear deeper in the \gls{bb} tree \citep{Linderoth1999} but neglects to consider improvements to the dual bound.
For pure feasibility problems, such as \gls{sat}, it is the preferred strategy \citep{Biere2015,Achterberg2007}.

In \gls{mip}, a node's \gls{lp} subproblem differs only little from the one of its parent node.
{\Gls{dfs}} node selection hence allows for an efficient resolving of the subproblem \citep{Achterberg2007}.
A similar effect may appear for the \gls{lcsp} if a dynamic shortest path algorithm is used in \cref{lcsp:algo:bb:sp}. 

\emph{Most-feasible search} \citep{Wojtaszek2010} selects the node whose solution is closest to feasiblity.
It hence focuses on primal improvements as well.
In \gls{mip}, a node's infeasibility is calculated as the sum of fractional values in its \gls{lp} solution.
We adapt it for \cref{lcsp:algo:bb} by replacing fractionality with the number of violated clauses as the measure of infeasibility:
In \cref{lcsp:algo:bb:nodeselection}, we always choose an assignment $\trail\in Q$ with 
\begin{equation}
\trail \in \argmin_{\trail'\in Q} \viol_{\trail'}.
\end{equation}

In contrast, \emph{best-first search} aims to improve the global dual bound.
It always selects an unprocessed assignment $\trail \in Q $ minimizing the local dual bound, i.e., 
\begin{equation}
\trail \in \argmin_{\trail'\in Q} \weight\left(p_{\trail'}\right).
\end{equation}
For a fixed branching order, {best-first search} (with appropriate tie-breaking) minimizes the size of the search tree \citep{Achterberg2007}. 
\emph{Best-first search with plunging} aims to combine the advantages of {\gls{dfs}} and {best-first search}.
We select a child of the current node, or, if there is none, a sibling. If neither exists, {best-first search} is applied \citep{Achterberg2007}.

More sophisticated rules like \emph{best-projection search} (following \citet{Linderoth1999} due to
\citet{Hajian1995} and \citet{Hirst1969}) and \emph{best-estimate search} \citep{Benichou1971} 
 estimate the objective value of feasible solutions contained in the subtree at a candidate node \citep{Achterberg2007}.
{Best-estimate search} relies on the computation of \emph{pseudocosts}.
Pseudocosts are based on a measure of fractionality of individual variables in the \gls{lp} solution,
making {best-estimate search} inapplicable for \gls{lcsp}.

Instead, we adopt \emph{best-projection search}, which requires only a global measure of infeasibility.
As in {most-feasible search}, we measure the infeasibility of an assignment $\trail$ by $\viol_\trail$.
We select an assignment $\trail \in Q$ such that
\begin{equation} \label{lcsp:eq:proj}
 \trail\in \argmin_{\trail'\in Q}
    \left( 
        \weight\left(p_{\trail'}\right) + 
\left(\frac{\weight\left(p^*\right) - \weight\left(p_\emptyset\right)}{\viol_\emptyset}\right)\viol_{\trail'}\right),
\end{equation}
where $p(\emptyset)$ is the path computed for the root assignment and $\viol(\emptyset)$ the number of clauses violated by $p(\emptyset)$.
In \cref{lcsp:eq:proj}, the term 
$\left(\weight\left(p^*\right) - \weight\left(p_\emptyset\right)\right)/\viol_\emptyset$
represents an estimate of the change in the objective obtained by a unit change in infeasibility \citep{Linderoth1999}.

\subsection{MIP-Inspired Branching Rules}\label{lcsp:sec:mipbranching}

\Gls{mip} branching rules choose a variable among the fractional variables in the current \gls{lp} solution.
For the \gls{lcsp}, this corresponds to choosing an unassigned variable that is part of a conflict.
Classical rules are \emph{strong branching} \citep{Applegate1995} and \emph{pseudocost} branching \citep{Benichou1971} as well as combinations thereof, for example  \emph{reliability branching} \citep{Achterberg2007}.
The current state-of-the-art \citep{Gamrath2018,SCIP2024}, \emph{hybrid branching} \citep{Achterberg2009}, extends reliability branching with domain reduction rules and conflict-based variable scoring.
Pseudocost branching and its derivatives rely on a measure of fractionality of individual variables in an \gls{lp} solution.
They are therefore not applicable for the \gls{lcsp}. Instead, we will focus on strong branching.

Let $\trail$ be the current assignment and $\confl\subseteq \VS$ a conflict derived from \mbox{$\operatorname{SAT}(\form \cond \trail\cup\tent_{p_\trail})$}.
For each $\vs\in\confl$, we evaluate the \emph{up} branch $\trail\cup\{\vs\}$ and the \emph{down} branch $\trail \cup \{\neg \vs\}$
by performing unit propagation (\cref{lcsp:algo:bb:propagation}), enforcement (\cref{lcsp:algo:bb:enforce}), and the shortest path search (\cref{lcsp:algo:bb:sp}).
For each branch $\trail \cup \{\lit\}$ with $\lit\in\{\vs,\neg \vs\}$, we then either find a path $p_{\trail\cup\{\lit\}}$
or derive infeasibility.
We let  $\gamma_\lit : = \weight(p_{\trail\cup \{\lit\}}) - \weight(p_\trail)$ with $\weight(p_{\trail\cup \{\lit\}}) := \infty$ in case of infeasibility and
use the product rule \citep{Achterberg2007} to calculate the variable score
\begin{equation}\label{lcsp:eq:score}
    \operatorname{score}\left( \vs \right) := \max \left(\epsilon, \gamma_{\vs} \right) \max\left(\epsilon, \gamma_{ \neg\vs} \right).
\end{equation}
The parameter $\epsilon >0$ avoids the score collapsing to zero if no improvement was made in the up or down branch.
We select a branching variable $\vs\in\confl$ with
\begin{equation}
    \operatorname{score}\left(\vs\right) = \min_{\vs'\in\confl}\left(\operatorname{score}\left( \vs' \right) \right).
\end{equation}
%
%

Full strong branching results in small search trees at the expense of computing additional subproblems, usually resulting in a slower overall search \citep{Santanu2024}.
Therefore,  \emph{strong branching with working limits} \citep{Achterberg2007} stops evaluating variables if no improvement has been made after $ L \left(1-\xi \right) $ evaluations \citep{SCIP2024, Mexi2024}.
Here, $L$ is called the look-ahead parameter, and $\xi$ is the fraction of variables in $\confl$ that have not yet been considered.
\Gls{mip} solvers usually also limit the number of Simplex iterations \citep{Achterberg2007}, which is not applicable here.

\subsection{SAT-Inspired Branching Rules}\label{lcsp:sec:satbranching}

Nowadays, \gls{sat} variable selection appears to be studied exclusively for 
\gls{cdcl} solvers, where state-of-the-art rules are all based on learned clauses \citep{Biere2015}.
These rules are often derived from the \gls{vsids} rule \citep{Moskewicz2001,Biere2015}.
\Gls{vsids} keeps a score for each variable and branches on variables with the highest score.
When a conflict is found, \gls{cdcl} solvers learn a new clause 
that is usually far smaller than the current assignment via an implication graph. 
The score of variables in this clause is then incremented (\emph{bumped}).
Scores are initialized with the number of occurrences of a variable in clauses.
In regular intervals, they are multiplied with a decay factor. This avoids branching on variables that are no longer significant in the current region of the search tree.
We cannot employ {\gls{vsids}} out of the box, as learning clauses would involve repeated resolves of the shortest path problem, which is computationally too expensive.
Note also that, in \gls{sat}, the conflict is found in the current assignment.
In \cref{lcsp:algo:bb}, a conflict is caused by the shortest path solution and involves variables that are \emph{not} in the assignment.
We adopt {\gls{vsids}} as follows: When the empty clause is derived  in \cref{lcsp:algo:bb:propagation}, we bump all assigned variables in $\trail$.
When the  assignment $\tent$ induced by the shortest path in \cref{lcsp:algo:bb:tent} results in an unsatisfiable formula $\form \cond \tent$, we bump all variables in  $\trail$ and in the conflict.
We call this modified rule \gls{cvds}. 

Branching rules have also been studied for 
\gls{dpll} solvers \citep{DPLL1962}. For an overview, see \citep{Lagoudakis2001}. 
We adopt two rules:
The simple {\gls{moms}} rule \citep{Lagoudakis2001} branches on the variable that appears the most in clauses of minimum size.
Letting $k := \min_{\clause \in \form\cond\trail} |\clause|$,
it selects a variable $\vs\in\confl$ that maximizes
\begin{equation}\label{lcsp:eq:moms}
 \left|\left\{ \clause \in \form\cond\trail: |\clause| = k \land \left(\vs\in\clause \lor \neg\vs\in\clause\right)\right\}\right|.
\end{equation}
\Cref{lcsp:eq:moms} roughly approximates the strength of the unit propagation triggered by $\vs$.

The {\gls{up}} rule \citep{Crawford1996,Li1997}, in contrast, explicitly computes the number of unit propagations for each $\vg\in\confl$.
The score of a variable is calculated following \cref{lcsp:eq:score} with $\gamma_{\vs},\gamma_{\neg\vs}$ set to the number of propagations in the up and down branch, respectively.
If infeasibility is detected in the up (down) branch, we let $\gamma_{\vs} :=\infty$ ($\gamma_{\neg \vs} :=\infty$).
For the \gls{lcsp}, {\gls{up}} can be seen as a relaxation of strong branching.
The rationale is that unit propagation leads to deleted arcs in $\G$, which will in turn drive up the cost of a shortest path.

Two variants of the unit propagation rule are conceivable for \cref{lcsp:algo:bb}:
Given an assignment $\trail$ and a candidate variable $\vs\in\confl$, {\gls{sup}} computes
$\trail' := \FuncSty{propagate}(\form\cond \trail \cup \{ \lit \})$ for $\lit\in \{\vs,\neg\vs\}$
and sets 
\begin{equation}
    \gamma_{\lit} := \left| \trail' \right| - \left|\trail\cup\{\lit\}\right|.
\end{equation}
In contrast, {\gls{dup}} computes $\trail'$ by running the full propagation and enforcement loop in
\crefrange{lcsp:algo:bb:propEnforceLoop}{lcsp:algo:bb:propEnforceLoopEnd}, i.e, it also considers additional propagations that are caused by enforcements in the graph.
In {\gls{sup}}, infeasibility can be detected if the empty clause is derived during unit propagation.
In {\gls{dup}}, infeasibility can additionally be found by the check in \cref{lcsp:algo:bb:enforceinfeas}.

{\Gls{up}} is considered computationally expensive as modern \gls{sat} solvers spend most of their run time on unit propagation \citep{Lagoudakis2001}. 
Due to the invocation of a shortest path search in every node, this assessment need no longer hold for the \gls{lcsp}, however.

\section{Application: Flight Planning With Traffic Flow Restrictions}\label{lcsp:sec:tfr}
%
A \emph{projected airway network} $\twodgraph$  is a directed graph representing the two-dimensional projection of a $3$-dimensional airway network $\threedgraph$. Vertices in $\twodgraph$ correspond to coordinates on Earth.
The distance between two vertices is well-defined as the \emph{\gls{gcd}} between them. 
In flight planning, the distance between the departure and the destination airport is usually not an objective. 
However, commonly used objectives, such as the fuel consumption or the duration, correlate with the flight's distance.

Routing is performed on $\threedgraph$. Due to its prohibitively large size, $\threedgraph$ is only stored  implicitly 
as a copy of $\twodgraph$ in every \emph{flight level}.
A flight level $\ell\in L$ is an altitude $\altitude{\ell}$ at which commercial aircraft are allowed to \emph{cruise} between vertices that are adjacent in $\twodgraph$. 

For \emph{climbing} and \emph{descending} from vertex $u$ at level $\ell$ to vertex $v$ at level $\ell'$ it is not only necessary that $u$ and $v$ are adjacent in $\twodgraph$. The speed and climb rate of the considered aircraft need to be such that the altitude difference $\altdiff_{\ell, \ell'}$ is \emph{flyable} in less than the distance between $u$ and $v$ in $\twodgraph$ (cf. \cref{lcsp:eq:levelChangeAllowed} in \cref{lcsp:appendix:pedric}).

The costs of the cruise, climb, and descend maneuvers depend on the aircraft weight \citep{Blanco2022}, the weather conditions \citep{Blanco2016a}, and overflight costs \citep{Blanco2016, Blanco2017}, all of which are not relevant in the \gls{lcsp} setting since \glspl{tfr} do not depend on them. 
In this chapter, we thus work with simplified, easy-to-model cost functions (cf. \cref{lcsp:eq:duration,lcsp:eq:consumption} in \cref{lcsp:appendix:pedric}) to guarantee 
that the results are easy to reproduce.

Traffic flow restrictions (\glspl{tfr}) are logical constraints imposed on $\threedgraph$. 
Clearly, the calculation of cost minimal routes between two airports in $\threedgraph$ s.t.\ no \gls{tfr} is violated gives rise to an \gls{lcsp} instance (cf. \Cref{lcsp:sec:introduction}).

\glspl{tfr} are given as a conjunction of \emph{restrictions}. Each restriction
is in \gls{dnf}, i.e., it is a disjunction of conjunctive \emph{clauses}.
Literals in these clauses correspond to arrival or departure events, arcs, or vertices at specific height intervals. 
They don't correspond to arcs in $\threedgraph$, but instead to sets of arcs and also vertices. Adapting \cref{lcsp:algo:bb} is straightforward.

We transform the traffic flow restrictions $\tfrform$ over the graph variables $\VG$ into a \gls{cnf} formula $\form$ using the standard Tseitin transformation \citep{Tseitin1983}.  
As discussed in \cref{lcsp:sec:problem_formulation}, this leads to the introduction of free variables $\VF$ that do not correspond to arcs or vertices in $\G$.
The transformation is performed as follows: We begin with an empty \gls{cnf} formula $\form$.
For each restriction $R$ in $\tfrform$, which can be written as
\begin{equation}
R = ( \lit_{11} \land \dots \land \lit_{1k_1} ) \lor \dots \lor ( \lit_{r1} \land \dots \land \lit_{r k_r} ),
\end{equation} we introduce free variables $C_1,\dots C_r$ to $\VF$, 
add 
 $C_1\lor \dots \lor C_r$ to $\form$, and, for each $i \in 1,\dots, r$, we add the clauses
\begin{equation}\label{lcsp:eq:equisat}
C_i \lor \neg \lit_{i1} \lor \dots \lor \neg \lit_{ik_i} 
\text{ and }
\bigwedge\limits_{j\in 1,\dots,k_i}  \neg C_i \lor \lit_{ij}    
\end{equation}  
via conjunction to $\form$.
\Cref{lcsp:eq:equisat} implies 
\begin{equation}\label{lcsp:eq:clausetovar}
C_i \Leftrightarrow (\lit_{i1} \land \dots \land \lit_{ik_i} )    \quad \quad \forall i\in 1,\dots,r,
\end{equation}
i.e, we can identify the \gls{dnf} clause $\lit_{i1} \land \dots \land \lit_{ik_i} $ with the variable $C_i$ 
and write $\lit_{ij}\in C_i$ for any $\lit_{ij}$ with $j\in1,\dots,k_i$.

Applying this procedure yields a \gls{cnf} formula $\form$ that is equisatisfiable to $\tfrform$ but 
not equivalent (due to the introduction of new variables).
Still, any complete assignment that satisfies $\tfrform$ trivially induces a (partial) assignment that satisfies $\form$ and that, by \cref{lcsp:eq:equisat}, can be transformed into a complete assignment in linear time.
In particular, this allows us to uniquely extend any assignment $\tent := \trail \cup \tent_p$ induced by $p$ to a complete assignment.
Checking satisfiability of $\form\cond\tent$ in \cref{lcsp:algo:bb:satresolution} then reduces to checking whether $\form\cond\tent$ is the empty set.

If $\form\cond\tent$ is not empty,
we can thus easily derive the conflict $\confl$ by listing all clauses in $\form \cond \trail$ that $\tent$ does not satisfy, i.e,
\begin{equation}\label{lcsp:eq:conflict}
\confl := \{ \vg : \vg \in \clause \text{ or } \neg \vg \in \clause \text{ for some } \clause \in \form \cond \trail \text{ with } \{\alpha\} \cond \tent \neq \emptyset \}.
\end{equation}



 Recall that by \cref{lcsp:prop:disjointbranching} only branching on graph variables guarantees that no path is repeated, which is why we opt for expressing conflicts in terms of graph variables only.
By \cref{lcsp:eq:clausetovar}, it is straightforward to obtain the graph conflict
\begin{equation}\label{lcsp:eq:graphconflict}
    \confl_{\cond\VG} := (\confl \cap \VG) \cup \{ \vg \in \vf : \vf \in \confl\cap \VF \}.
\end{equation}


\subsection{Dynamic Shortest Path Search}\label{lcsp:sec:dsp}
During each iteration $i$ of \cref{lcsp:algo:bb}, we solve the shortest path instance $(\twodgraph_{\cond\trail_i}, s, t)$ defined by the assignment $T_i$.
We exploit the similarity of the graphs $\twodgraph_{\cond\trail_i}$  and $\twodgraph_{\cond\trail_{i+1}}$ by using a  variant of the exact dynamic shortest path algorithm \gls{lpa}  \citep{Koenig2004}.
\Gls{lpa} combines $\text{A}^\star$ search \citep{Hart1968} with ideas from the Dynamic-SWSF-FP algorithm \citep{Ramalingam1996b}.


\color{black}
In the following, we concisely recap the algorithm and tailor it to the specific structure of  airway networks.
For a detailed proof and pseudocode, see \citet{Koenig2004}.

For the \gls{fpp} with real aircraft performance functions, appropriate heuristics for $\text{A}^*$ are available in the literature \citep{Blanco2022}. 
Using our simplified aircraft model, we define the heuristic $h(\threedvrt)$ for a vertex $\threedvrt\in \V(\threedgraph)$ as an underestimator of the fuel consumption of reaching the target from $\threedvrt$.
It is calculated based on the great circle distance between $\threedvrt$ and the target, assuming the optimal flight level.

\Gls{lpa} maintains two labels for each vertex in $\threedvrt \in V(\threedgraph)$ ,
 a \emph{distance estimate} $d(\threedvrt)$ and a \emph{look-ahead} estimate $\rhs(\threedvrt) := \argmin_{ \threedvrtalt \in \ingoing{\threedvrt}}d(\threedvrtalt)+\weight_{\threedvrtalt\threedvrt}$.
A vertex $\threedvrt \in V(\threedgraph)$ is \emph{inconsistent} if $d(\threedvrt) \neq \rhs(\threedvrt)$. 
An inconsistent vertex is \emph{overconsistent} if $d(\threedvrt) > \rhs(\threedvrt)$ and \emph{underconsistent} otherwise.
\Gls{lpa} maintains a priority queue of inconsistent vertices $\threedvrt$ ordered by $\min(d(\threedvrt), \rhs(\threedvrt)) + h(\threedvrt)$.

Let $\trail_0,\dots,\trail_r$ be the sequence of assignments in order of their extraction from the queue in \cref{lcsp:algo:bb:deque} of \cref{lcsp:algo:bb}.
The invocation of \gls{lpa} on $\threedgraph_{\cond\trail_0}$ is equivalent to a standard $A^\star$ search.

Each subsequent search on $\threedgraph_{\cond\trail_i}, 0 < i \leq r$ then begins with an initialization phase.
First, we record the symmetric difference $\mathcal{D}$ of $A(\threedgraph_{\cond\trail_{i-1}})$ and $A(\threedgraph_{\cond\trail_i})$.
For each arc $(\threedvrtalt,\threedvrt)\in \mathcal{D}$,
we recompute the value $\rhs(\threedvrt)$.
All pairs that become inconsistent by this operation are added to the priority queue (of \gls{lpa}).

During the main phase of \gls{lpa}, inconsistent vertices $\threedvrt$ are extracted
from this priority queue.
If $\threedvrt$ is overconsistent, its distance label $d(\threedvrt)$ is set to $\rhs(\threedvrt)$.
If it is underconsistent, it is set to $\infty$. 
Then, $\rhs(\threedvrtalt)$ is updated for all $\threedvrtalt$ in the out-neighborhood of $\threedvrt$. 
If $\threedvrt$ was overconsistent, we only need to compute $\min(\rhs(\threedvrtalt), d(\threedvrt)+\weight_{\threedvrt\threedvrtalt})$
to obtain the new value of $\rhs(\threedvrtalt)$.
Otherwise, however, the recomputation requires the full iteration over the in-neighborhood of $\threedvrtalt$.

The graph $\threedgraph$ is characterized by large neighborhoods:
The in-neighborhood $\ingoing{\threedvrtalt}$ of  $\threedvrtalt = (u,\ell)\in \V(\threedgraph)$
is of size $|L| \ingoing{u}$. 
In the data set used for the computational results in \cref{lcsp:sec:computation}, there are $|L| = 181$ flight levels.
Moreover, there is no guarantee that the distance labels of vertices in $\ingoing{\threedvrtalt}$
are correct when $\rhs(\threedvrtalt)$ is calculated and $\rhs(\threedvrtalt)$ might be recomputed many times before it attains its correct value.

We avoid this issue by modifying a trick that \citet{Bauer2009} adapted from \citet{Narvaez2000} for the Dynamic-SWSF-FP algorithm by \citet{Ramalingam1996b}.
Let $\mathcal{B}$ be the shortest path tree calculated in $\threedgraph_{\cond\trail_{i-1}}$ in the $i-1$'th invocation of  \gls{lpa}.
In the initialization phase, for any deleted arc $(\threedvrtalt,\threedvrt) \in \A(\threedgraph_{\cond\trail_{i-1}})\backslash \A(\threedgraph_{\cond\trail_i})$
with $(\threedvrtalt,\threedvrt) \in \mathcal{B}$, we identify the subtree $\mathcal{B}'$ of $\mathcal{B}$ rooted at $\threedvrt$.
For each $\threedvrtalt \in \mathcal{B}'$, we set the distance label $d(\threedvrtalt)$ to infinity.
In a second step, we recompute $\rhs(\threedvrt)$ for all $\threedvrt$ for which $d(\threedvrt)$ was changed.
In contrast to \citet{Bauer2009}, we do not propagate inserted arcs along the search tree.
With these changes, each vertex's in-neighborhood is iterated at most once, namely in the initialization phase, and in the main phase, only overconsistent vertices are encountered.
We refer to this variant of the algorithm as \gls{lpan}.
%

\section{Computational Experiments}\label{lcsp:sec:computation}

In the following, we evaluate the performance of \cref{lcsp:algo:bb} on a real-world airway network and \gls{tfr} system obtained from 
\gls{lhsy}. \Cref{lcsp:algo:bb} has four degrees of freedom: the conflict, the node selection rule, the branching rule and the shortest path algorithm.
In three experiments, we benchmark the impact of their various configurations introduced in \cref{lcsp:sec:conflict,lcsp:sec:nodeandbranch,lcsp:sec:dsp}.

All input data and experimental results from this section are, in anonymized form, publicly available in the supplementary material \citep{Code2024}.
Real aircraft performance functions are confidential and not part of the dataset. 
Instead, we use the easily reproducable, artificial aircraft model described in \cref{lcsp:appendix:pedric}.
The aircraft model is specifially designed to produce realistic trajectories that touch the same \glspl{tfr} as 
real aircraft trajectories.
Contrary to real-world applications, the  fuel consumption of the artificial aircraft is not time-dependent.
This ignores the huge impact of weather  on an aircraft's fuel consumption. However, we do not have access to weather prognosis and, most importantly, those would not add value to our contribution: the design of an algorithm for the LCSP, exploiting the interplay of the shortest path and the logic layers.
Additionally, this allows us to benchmark \cref{lcsp:algo:bb} against the \gls{mip} formulation \tfrmip on small instances. Adding time-dependency to the \gls{mip} is possible \citep[cf.][]{Nannicini2010} but the model size explodes already for small instances.

\subsection{Input Data}

Our $2$d airway network $\twodgraph$ has $\num{138923}$ vertices and $\num{962145}$ arcs covering the whole globe. 
Together with $181$ available flight levels at different altitudes, the $2$d airway network defines an implicit $3$d airway network $\threedgraph$ with roughly $25$ million vertices and $51$ billion arcs.
Due to the network's large size, we only keep $\twodgraph$ in storage and create only necessary parts of $\threedgraph$ on the fly.

The \gls{tfr} system consists of \num{18238}  \glspl{tfr} extracted from the systems of \gls{lhsy}, that were active at some point during the 24 hours after February 22, 2022, 22:00, which is the departure time of all flights in our experiments. 
The latest \glspl{tfr} are also published by \citet{Eurocontrol}.

For each \gls{od}, we compute a fuel consumption minimal trajectory in a subgraph  of $\threedgraph$ in which every vertex fulfills $\mathtt{gcd}(s,v) + \mathtt{gcd}(v,t) \leq 1.2\mathtt{gcd}(s,t)$.
All \glspl{tfr} outside the resulting search space are dropped.
Per \gls{od}, we thus considered a \gls{tfr} system with on average \num{4615} restrictions on \num{9032} variables.
After assigning arrival and departure variables, we applied the Tseitin transformation, resulting in a \gls{cnf} formula with \num{15533} variables and \num{30964} clauses on average.

\subsection{\texttt{PedRicAir}, a Naive Aircraft Performance Model} \label{lcsp:appendix:pedric}

\begin{figure}[t]
    \centering
    \includegraphics[]{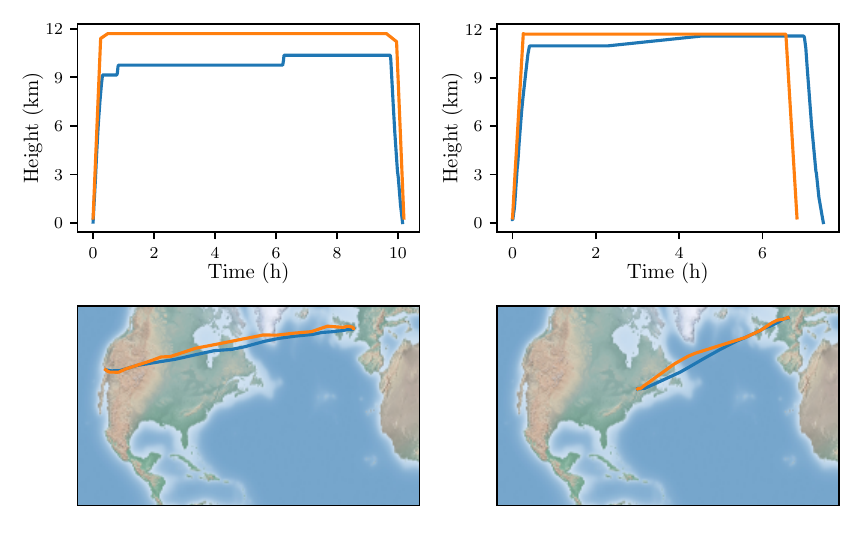}
    \caption[Comparison between real and artificial aircraft]{Height profile and skypath of PedricAir (orange) compared to two historic trajectories flown with real aircraft (blue). Left: American Airlines Flight 137  from London Heathrow Airport to Los Angeles International Airport on March 27th 2022.
    Right: Lufthansa flight 478  from Frankfurt Airport to Montréal Pierre Elliott Trudeau International Airport on March 30th 2022.
    Flight data courtesy to FlightRadar24.
    }
    \label{lcsp:fig:pedricexample}
\end{figure}
In commercial aviation, arc costs are usually calculated using so-called aircraft performance tables \citep[cf.][]{Blanco2023}.
These are, however, a well-kept industry secret.
To make the computational results in this section reproducible, a simplified aircraft model called \texttt{PedRicAir} is employed%
that is described in the following.
The model is fairly basic, but specifically designed to maintain realistic horizontal and vertical flight profiles. 
Thus, the set of relevant \glspl{tfr} for \texttt{PedRicAir}'s trajectories is similar to the set of \glspl{tfr} relevant in practice.
In \cref{lcsp:fig:pedricexample}, we compare some trajectories obtained using \texttt{PedricAir} with those
flown by real aircraft in the past.
\begin{figure}
    \begin{subfigure}{.49\linewidth}
    \centering
        \begin{tikzpicture}[
        scale=1,
        font={\fontsize{9pt}{10}\selectfont},
        arc/.style={-latex, thick}]
            \node[circle] (source) at (0,0) {};
            \node[circle, above right = 3.5cm and 3.5cm of source] (TOC) {};                 
            \node[circle] (target) at (source -| TOC) {};
            
            \draw[-] (source) -- (TOC) node[midway, sloped, fill=white] {$\speed = 240.1 \frac{m}{s}$};
            \draw[-] (TOC) -- (target) node[midway, fill=white] {$\climbrate = 12.7 \frac{m}{s}$};
            \draw[-] (source) -- (target) node[midway, fill=white] {$\projspeed = \sqrt{\speed^2 - \climbrate^2}$};
            \node[] at (-.5,-1) {};
    \end{tikzpicture}
    \end{subfigure}
    \begin{subfigure}{.49\linewidth}
    \centering
        \begin{tikzpicture}[scale=1,
        font={\fontsize{9pt}{10}\selectfont},
        arc/.style={-latex, thick}]

        \node[circle] (source) at (0,0) {$(u,\ell)$};
        \node[circle, below = .5cm of source] (projectedSource) {$(u,0)$};
        
        \node[circle, above right = 1.5cm and 1.5cm of source] (TOC) {$\mathtt{TOC}$}; 
        
        \node[circle, right = 1.5cm of TOC] (target) {$(v,\ell')$};
        
        \node[circle] (projectedTarget) at (projectedSource -| target) {$(v,0)$};
        
        \draw[-] (source) -- (TOC);
        \draw[->] (TOC) -- (target);
        
        \draw[->, dashed] (projectedSource) -- (projectedTarget);
        
        \node[] (projectedTOC) at (projectedSource -| TOC) {};
        
        \path (projectedSource.north west)
        edge[decorate,decoration={brace,raise=.15cm},"$d_{\mathtt{climb}(\ell, \ell')}$"above=6pt]
        (projectedSource.north west -| projectedTOC.north east);
    %
        
        \path (projectedSource.south west)
        edge[decorate,decoration={brace,mirror,raise=.15cm},"$d_{uv}$"below=6pt]
        (projectedSource.south west -| projectedTarget.south east);
        \end{tikzpicture}   
    \end{subfigure}
    \caption[Speed characteristics of the artificial aircraft]{
Left: The speed triangle of the artificial aircraft.  
When climbing along an arc, the projected speed $\projspeed$ is required to calculate the \emph{\gls{toc}} point. 
Right: Typical climb maneuver of the artificial aircraft. Depicted is a climb maneuver from
a level $\ell$ to a higher level $\ell'$ along an arc $(u,v)\in\twodgraph$ with length $d_{uv}$.
After reaching \gls{toc}, a cruise phase is entered until $v$ is reached.}\label{lcsp:fig:climb}
\end{figure}

\paragraph{Cruising}\texttt{PedRicAir} cruises at a constant speed of Mach $0.7$, which is equivalent to $\speed = 240.1 \frac{m}{s}$,
 independent of the flight level. 
For an arc $a\in A(\twodgraph)$ of length $d_a$ (in meters) at level $\ell$ the flight duration is hence $\frac{d_a}{\speed} \unit{s}$.


Aircraft typically have an optimal cruise altitude at which fuel consumption is minimal.
For \texttt{PedRicAir}, we define flight level $\ell^* := 181$, at altitude  \mbox{$\mathrm{alt}(\ell^*) = \num{11300}\unit{m}$}, to be optimal.
\citet{Kuehn2024}  specifies the typical consumption of a commercial aircraft as $0.03 {\frac{\mathrm{kg}}{\mathrm{km}}}$ per seat. 
For \texttt{PedRicAir}, we assume a capacity of $200$, resulting in a fuel consumption parameter $f:=6\frac{\mathrm{kg}}{\mathrm{km}}$. 
For all flight levels $\ell$, the actual fuel consumption is based on $f$ but increases as a function of the deviation in altitude from 
$\ell^*$.
Specifically, given a flight level $\ell$ at altitude $\altitude{\ell}$ and an arc $a$ with length $d_a$, the cruise consumption of \texttt{PedricAir} along $a$ is given by
%
%
\begin{equation}
    \label{lcsp:eq:cruiseconsumption}
    \cruiseconsumption{d_a,\altitude{\ell}} := \frac{d_a}{10^3} \cdot f \cdot 1.01^{1 + {|\altitude{\ell}-\altitude{\ell^*}|}/{500}} \unit{kg}.
\end{equation}

\paragraph{Climb and Descent} 
The climb and descent rates of the aircraft model are identical and fixed to  $\climbrate = 2.500\unit{\frac{ft}{min}} = 12.7\unit{\frac{m}{s}}$. 
\texttt{PedRicAir} is only allowed to perform \emph{step climbs}, i.e., continuous climb and descent manoeuvres cannot extend across multiple arcs. 
When reaching an arc's head, the aircraft is forced to be at a flight level.

Whether a step climb is possible, is calculated as follows:
Let \mbox{$(u,v)\in\A(\twodgraph)$} be the arc along which \texttt{PedRicAir} attempts a climb from level $\ell$ at $u$ to level $\ell'$ at $v$. 
We denote the difference in altitude between two flight levels by $\altdiff_{\ell, \ell'}:=|\altitude{\ell}-\altitude{\ell'}|$.
Due to the constant climb rate, the climb will be completed after exactly $\frac{\altdiff_{\ell, \ell'}}{\climbrate}$ seconds.
During this time, the airplane continues traveling  horizontally along $(u,v)$ with a projected speed of 
$ \projspeed := \sqrt{\speed^2 - \climbrate^2}$ (see \cref{lcsp:fig:climb}).
Hence, the airplane reaches $v$ after ${d_{uv}}/{\projspeed}$ seconds and the
climb is only permitted if
\begin{equation}
    \label{lcsp:eq:levelChangeAllowed}
    \frac{\altdiff_{\ell, \ell'}}{\climbrate} \leq \frac{d_{uv}}{\projspeed}.
\end{equation}
%
%
%
%
%
%
If the above inequality is not tight, the aircraft reaches its \emph{\gls{toc}} before reaching $v$. 
In the surface projection,
$d_{\mathtt{climb}(\ell, \ell')} := \projspeed\frac{\altdiff_{\ell, \ell'}}{\climbrate}$ meters along $(u,v)$ have been covered at this point.
To reach $v$ at level $\ell'$, the aircraft now cruises the remaining $d_{uv} - d_{\mathtt{climb}(\ell, \ell')}$ meters.
An example is depicted in \cref{lcsp:fig:climb}. 
For step descents, the calculations are done analogously.

We set the fuel consumption for a climb from $\ell$ to $\ell'$ that covers $d$ meters in the surface projection to be equivalent to 
cruising $d$ meters at altitude $\frac{\altitude{\ell} + \altitude{\ell'}}{2}$.
For step descents from $\ell$ to $\ell'$, we assume the consumption to be equivalent to cruising at the source level $\ell$.

Putting everything together, the flight time along an arc \mbox{$a=(u,v)\in\A(\twodgraph)$} between flight levels $\ell$ and $\ell'$
can be calculated as
\begin{equation}
    \duration(a,\ell,\ell') :=  \begin{cases}
        \frac{\altdiff_{\ell, \ell'}}{\climbrate} + \frac{d_a - d_{\mathrm{climb}(\ell,\ell')}}{\speed}
%
        & \text{ if } \frac{\altdiff_{\ell, \ell'}}{\climbrate} \leq \frac{d_a}{\projspeed}  \\
        \infty               & \text { otherwise.}
    \end{cases}\label{lcsp:eq:duration}
\end{equation}
If $\ell=\ell'$, we have $\altdiff_{\ell,\ell'} = 0$ and \cref{lcsp:eq:duration} reduces to the cruise case.
The fuel consumption is given by
\begin{equation}
\consumption(a,\ell,\ell') := 
\begin{cases}
     \cruiseconsumption{d_a,\altitude{\ell}}          & \text{ if } \ell = \ell' \\
     \cruiseconsumption{d_a,\altitude{\ell}}          & \text{ if } \ell > \ell' \text{ and } \frac{\altdiff_{\ell, \ell'}}{\climbrate} \leq \frac{d_a}{\projspeed}  \\
     \cruiseconsumption{d_a - d_{\mathrm{climb}(\ell,\ell')},\altitude{\ell'}} + \\
     \cruiseconsumption{d_{\mathrm{climb}(\ell,\ell')},\frac{\altitude{\ell}+\altitude{\ell'}}{2}}  &\text{ if } \ell < \ell' \text{ and } \frac{\altdiff_{\ell, \ell'}}{\climbrate} \leq \frac{d_a}{\projspeed}  \\
 \infty & \text{ otherwise.}
\end{cases}
\label{lcsp:eq:consumption}
\end{equation}


%

\subsection{Experiments}\label{lcsp:sec:experiments}

In total, we conduct four experiments.
The first three experiments benchmark the performance of various configurations of the conflict,
node selection rule and branching rule in \cref{lcsp:algo:bb}.
Additionally, the overall impact of choosing an appropriate configuration of these four degrees of freedom is evaluated by comparing against a naive, ad-hoc configuration
 which we refer to as the \emph{baseline}.


The baseline employs \emph{\gls{dfs}} node selection, which is the standard in \gls{sat} solving, together with an uninformed ad-hoc branching rule that simply branches on the first literal in the smallest violated clause (\emph{clause}). It hence branches on regular conflicts.
It also solves each \gls{sp} subproblem from scratch using standard $\mathrm{A}^*$ search.


In the first experiment, we fix \emph{\gls{sup}} branching on graph variables as the branching rule and benchmark different node selection rules against each other.
We evaluate the \emph{\gls{dfs}}, \emph{most-feasible search}, \emph{best-first search}, 
\emph{best-projection search}, and the \emph{best-first search with plunging} node selections rules.

In the second experiment, we fix the \emph{best-first search} node selection rule and benchmark different branching strategies and conflict choices. We evaluate the \emph{\gls{moms}}, \emph{strong branching}, \emph{\gls{cvds}}, \emph{\gls{sup}}, \emph{\gls{dup}}, and \emph{clause} branching rules. 
We applied \emph{\gls{moms}}, \emph{\gls{cvds}}, \emph{\gls{sup}}, and \emph{\gls{dup}} on regular conflicts  \eqref{lcsp:eq:conflict} as well as on graph conflicts \eqref{lcsp:eq:graphconflict}.
\emph{Strong branching} was applied with working limits and only on graph conflicts, as it performed poorly on regular conflicts in preliminary experiments. 
The choice of the fixed branching rule in the first experiment and the fixed node selection rule in the second experiment is motivated by the respective methods' strong performance in preliminary experiments. For the same reason, both experiments were conducted using \gls{lpan} as the shortest path algorithm.

In the third experiment, the effect of employing a dynamic shortest path algorithm is evaluated.
We compare \gls{lpa} and \gls{lpan} against a straightforward $\mathrm{A}^*$ search that recomputes every \gls{sp} query from scratch.
In all cases apart from the baseline, we use the best node selection and branching rules identified in the first two experiments.
Using {\gls{dfs}} for node selection results in subsequent \gls{sp} queries being closer to each other in the \gls{bb} tree \citep{Achterberg2005}.
We conduct an additional run of \gls{lpa} with \emph{\gls{dfs}} node selection to determine whether this leads to more similar search graphs that can be 
exploited by the \gls{lpa} algorithm.

In the fourth and final experiment, we compare the \gls{mip} formulation \tfrmip{}  against \cref{lcsp:algo:bb} 
using both the baseline configuration and the best overall configuration identified in the preceeding experiments, 
namely best-first node selection with \emph{\gls{sup}} branching on graph conflicts and \gls{lpan} shortest path search.

%
\Cref{lcsp:algo:bb} was implemented in C++20 and compiled with gcc 12.20. 
The \gls{mip} formulation was solved using the C++ interface of Gurobi 12.01 \citep{gurobi}.
All experiments were run on workstations with a \mbox{2 Ghz} Intel(R) Xeon(R) Gold 6338 CPU. 
Each invocation of \cref{lcsp:algo:bb} was allotted 120 GB \gls{ram};
for solving \tfrmip, we allotted 200 GB \gls{ram} to Gurobi.

\subsection{Instances}

Our instance set is defined by the subset of all \glspl{od} between 466 large international airports \citep{Code2024}. 
From this set, we exclude 545 \glspl{od} that are not flyable in the airway network.
For example, landing in Innsbruck, Austria, or overflying the Himalaya are known to be challenging operations in flight planning. Addressing these is beyond the scope of this paper. 
In total, \num{216145} \glspl{od} remain.
The hardness of these instances varies greatly due to the uneven distribution of \glspl{tfr} across the globe (\cref{lcsp:fig:tfronworld}). 
A few instances require, for some branching rules, several hundred \gls{bb} nodes to even obtain a feasible solution, whereas others are not affected by \glspl{tfr} at all.
In fact, using the baseline configuration of \cref{lcsp:algo:bb} (see \cref{lcsp:sec:experiments}),  \num{81.47}\% of the instances are solved directly in the root node and \num{95.43}\% are solved in less than 10 nodes.
We consider these instances \emph{trivial} and drop them from further consideration. 


\begin{figure}[t]
    \centering
    \includegraphics[]{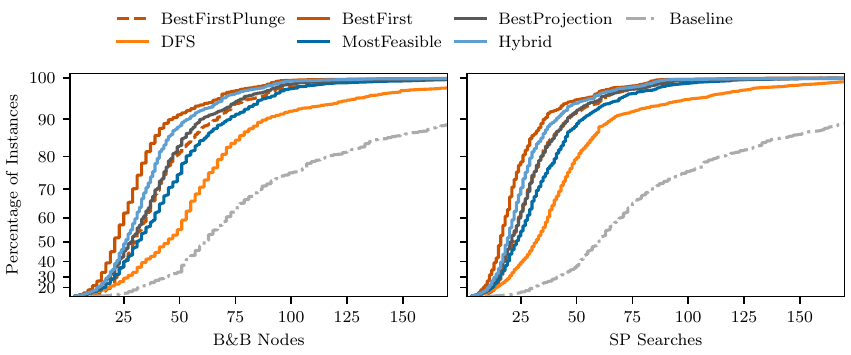}
    \caption[Cumulative frequency of B\&B nodes for node selection rules]{
        Cumulative frequency plot of solved instances per number of \gls{bb} nodes (left) and performed \gls{sp} queries (right) for the \emph{\gls{dfs}}, \emph{best-first search}, \emph{most-feasible search} and \emph{best-first search with plunging} node selection rules.
    The $x$-axis is truncated at 200 nodes, at which point 99.9\% of all instances are solved using \emph{best-first search} node selection.}
    \label{lcsp:fig:comp:nodeselection}
\end{figure}


This process leaves \num{9869} instances, on which we perform the first three experiments. 
In their analysis in \cref{lcsp:sec:results}, we filter out any instances that are consistently easy, that is, all instances that are solved in less than \num{50} nodes in \emph{all} tested combinations of branching and node selection rules, including the baseline.
We hence consider a final data set containing \num{2260} \glspl{od}, all of which admit an optimal solution.

\color{black}

In the last experiment, we compare \cref{lcsp:algo:bb} against the \gls{mip} formulation \tfrmip.
Due to the large size of the flight network $\threedgraph$, we only consider domestic instances, i.e.,
\glspl{od} for which the origin and destination airports are in the same country.
Extracting all such instances from the \num{9869} non-trivial instances yields a data set containing \num{106} \glspl{od}, 
all of which are in Europe.                                         

All results are available in the online appendix \citep{Code2024}.

\subsection{Results}\label{lcsp:sec:results}

Using a real aircraft model, the run time of \cref{lcsp:algo:bb} is heavily dominated by the \gls{sp} queries. 
This is because the computation of arc costs (weather- and weight-dependent duration and fuel consumption) is a complex task that cannot be precomputed. 
The most relevant performance measures to evaluate node selection and branching rules are thus the number of calls to the shortest path algorithm and the total number of performed arc relaxations across all \gls{sp} subproblems.
We provide these and additional performance measures, including the number of \gls{bb} nodes and run times, in \cref{lcsp:table:nodeselection,lcsp:table:branching,lcsp:table:hard}
in \ref{lcsp:appendix:results}. 
Even though we use the simplified \texttt{PedRicAir} aircraft model, the \gls{sp} search still accounts for more than 75\% of the run time in all configurations
of \cref{lcsp:algo:bb}.

\paragraph{Node Selection Rules}
For each node selection rule, we depict the cumulative frequency of solved instances over the number of \gls{bb} nodes and \gls{sp} searches in \cref{lcsp:fig:comp:nodeselection}.
All instances were solved to optimality, irrespective of the rule. 
%
\emph{Best-first search}, which aims at improvements in the dual bound, is the best node selection rule regarding all performance measures.
It requires {44.86}\% fewer B\&B nodes,  {43.85}\% fewer \gls{sp} searches, and {39.29}\% fewer arc relaxations 
than \emph{\gls{dfs}}, which performs worst.
After \emph{\gls{dfs}}, the second-worst performance (w.r.t. arc relaxations and \gls{sp} searches) is achieved by the \emph{most-feasible search} rule. 
Both of these rules aim at finding feasible solutions fast.

The remaining node selection rules \emph{hybrid search}, \emph{best-projection search}, and \emph{best-first search with plunging} all aim for a compromise between primal and dual improvements. This sophistication does not pay off: In \cref{lcsp:fig:comp:nodeselection}, we can see that, for all three rules, the cumulative frequency of solved instances
is contained in the band spanned by the \emph{most-feasible search} and \emph{best-first search} rules.
Moreover, among those three, the best-performing rule is \emph{hybrid search}, which is most similar to \emph{best-first search}.

\begin{figure}[t]
    \centering
    \includegraphics[]{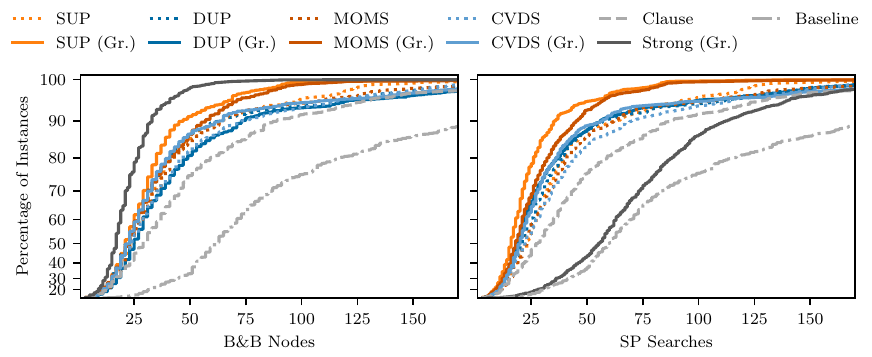}
     \caption[Cumulative frequency of B\&B nodes for branching rules]{
         Cumulative frequency plot of solved instances per number of B\&B nodes (left) and performed \gls{sp} queries (right) for different branching rules.
    The  $x$-axis is truncated at 200 nodes, at which point 99.9\% of all instances are solved using \emph{\gls{sup}} branching on graph conflicts. }
    \label{lcsp:fig:comp:branch}
\end{figure}

\paragraph{Branching Rules}
For each branching rule, we depict the cumulative frequency of solved instances over the number of \gls{bb} nodes and \gls{sp} searches in \cref{lcsp:fig:comp:branch}.
In all cases, all instances were solved to optimality.

The branching rule that attains the least number of \gls{bb} nodes is, unsurprisingly, \emph{strong branching}.
This is paid for by computing the most \gls{sp} queries, making it unsuitable for realistic flight planning applications.
Overall, the best-performing branching rule is \emph{\gls{sup}} on graph conflicts, which performs \num{43}\% fewer \gls{sp} searches and  \num{17.00}\% fewer arc relaxations than the baseline branching rule \emph{clause}.

The second-lowest number of \gls{sp} searches is attained by \emph{\gls{moms}} on graph conflicts, which requires \num{12.94}\% more searches than \emph{\gls{sup}}.
The lower computational effort of \emph{\gls{moms}} does not make up for this difference:
\emph{\gls{sup}} is still \num{9.50}\% faster, an effect that will be more pronounced for realistic airplane models,
and  uses \num{12.46}\% fewer arc relaxations.
\emph{\Gls{cvds}}, applied to graph or regular conflicts, outperforms the baseline branching rule \emph{clause}.
It is, however, surpassed by the simpler rules \emph{\gls{sup}} and \emph{\gls{moms}} that have been abandoned in the \gls{sat} community.
This may be due to the relatively small  \gls{bb} trees we observe, which disadvantage learning-based rules.
We note that we did not attempt to tune the parameters of \emph{\gls{cvds}}.

\begin{figure}[t]
    \centering
    \includegraphics[]{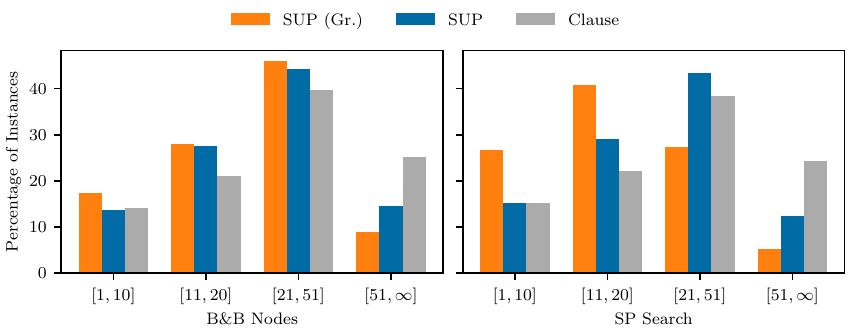}
    \caption[Comparison of graph and regular variable branching]{
        Comparison of the best branching rule \emph{\gls{sup}} on regular and graph conflicts using \emph{best-first search} node selection with  the \emph{clause} branching rule.
        We plot the percentage of instances solved depending on the number of \gls{bb} nodes (left) and on the \gls{sp} queries (right).
    }
    \label{lcsp:fig:comp:branchbars}
\end{figure}

\emph{\Gls{sup}} performs better than \emph{\gls{dup}} in both the number of \gls{sp} searches and arc relaxations.
This result surprises, given that \emph{\gls{sup}} is essentially a relaxation of \emph{\gls{dup}}.
We attribute this to the following reason:
\emph{\gls{sup}} performs a single round of unit propagation. 
This ensures that only unit propagations are counted that can be directly derived from enforcing/forbidding an arc in the current path.
In contrast,  \emph{\gls{dup}} performs multiple rounds of unit propagation and enforcement in the graph.
Enforcing an arc leads to the deletion of all arcs that could be used to bypass it w.r.t the topological sorting on $\RD$.
Such arcs may lie in parts of $\RD$ that are far away and never considered in the shortest path search.
Their deletion may consequently trigger unit propagations in irrelevant clauses that are not violated by any computed path.
In the \emph{\gls{dup}} rule, these unit propagations are counted towards a candidate variable's score and hence steer
the search away from more productive branching decisions.

\Cref{lcsp:prop:disjointbranching} suggests branching on graph conflicts.
Indeed, compared to branching on regular conflicts, it  reduces  
the number of \gls{sp} searches by between \num{6.51}\% (\emph{\gls{dup}}) and \num{30.11}\% (\emph{\gls{sup}}).
We illustrate this difference for \emph{SUP} in \cref{lcsp:fig:comp:branchbars}.

When branching on graph conflicts instead of regular conflicts, the number of arc relaxations per \gls{sp} search
increases by between  \num{16.23}\% (\emph{\gls{moms}}) and \num{35.25}\% (\emph{\gls{sup}}).
This behavior can be explained as follows:
First, branching on regular conflicts produces larger trees, and the size of a routing graph $\RD$
correlates negatively with its depth in the \gls{bb} tree.
Second, a regular conflict (\cref{lcsp:eq:conflict}) contains free variables that represent subformulae of a \gls{tfr}.
Assigning a truth value to such a subformula then leads to the assignment of multiple graph variables via unit propagation.
This consequently leads to more enforcements in the routing graph than if branching were done on graph conflicts.
As \emph{\gls{sup}} chooses a branching variable that causes the most unit propagations, the effect is most prominent for this rule.  
When branching on regular conflicts, \emph{\gls{cvds}} (\num{3.85}\%) and \emph{\gls{dup}} (\num{11.32}\%) require fewer arc relaxations in total, even though they produce larger \gls{bb} trees. This favors the second explanation.
%

%
%
The least  total number of arc relaxations is still attained by \emph{\gls{sup}} branching on graph conflicts:
Here, the effect of branching on graph conflicts on the number of \gls{sp} searches outweighs the increased effort 
for solving the \gls{sp} subproblem.

Combining the \emph{\gls{sup}} branching rule, the \emph{best-first search} node selection rule and
branching on graph conflicts appears to be the most promising configuration of \cref{lcsp:algo:bb}.
Compared to the baseline of \emph{\gls{dfs}} node selection with \emph{clause} branching,
it obtains an overall reduction of \num{75}\% in the number of \gls{sp} searches.

We also considered the \num{70} most difficult instances (see \cref{lcsp:table:hard}), that is, instances that require 
more than 300 \gls{bb} nodes for any configuration, including the baseline.
On this instance set, the number of \gls{sp} searches even decreases by \num{93}\% compared to the baseline.

\begin{figure}
    \centering
    \includegraphics[]{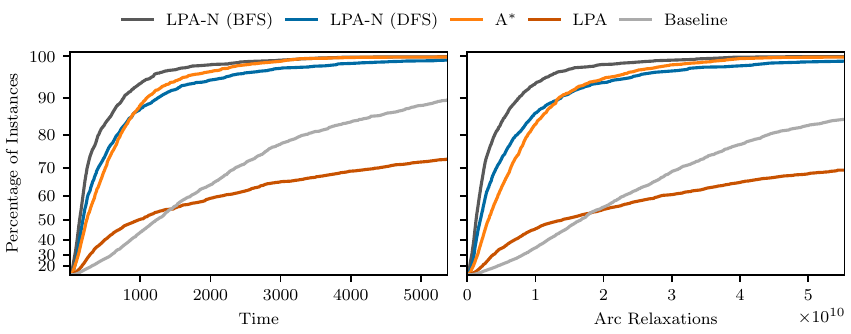}
    \caption[Comparison of shortest path algorithms]{
        Comparison of dynamic shortest path algorithms (\gls{lpa}, \gls{lpan}) with standard $\mathrm{A}^*$ search.
        We plot the percentage of instances solved depending on the time spent  (left) and on the number of arc relaxations performed (right).
        In both plots, the $x$-axis is truncated after \num{99.9}\% of the instances are solved by $\mathrm{A}^*$ search.
    }
    \label{lcsp:fig:comp:dsp_freq}
\end{figure}

\paragraph{Shortest Path Algorithm}
In the third experiment, we compared different variants of the \gls{lpa} algorithm with standard $\mathrm{A}^*$ search 
and the baseline.
In all cases, apart from the baseline, we employ \emph{\gls{sup}} branching on graph conflicts.
\emph{Best-first search} node selection is used for $\mathrm{A}^*$ and \gls{lpa}.
For \gls{lpan}, we test both \emph{best-first search} and \emph{\gls{dfs}} node selection.
The results are depicted as a cumulative frequency plot in \cref{lcsp:fig:comp:dsp_freq} (see also \cref{lcsp:table:dynamicsp}).

\Gls{lpa} solved all but one instances to optimality: The instance \emph{Brindisi --- Ontario} was terminated after reaching the time limit of 5 days.
This instance is omitted from the results presented in \cref{lcsp:fig:comp:dsp_freq,lcsp:table:dynamicsp}.
All other shortest path algorithms solved all instances to optimality.

\Gls{lpa} is, by a wide margin, not only outperformed by the standard $\mathrm{A}^*$ algorithm but even by 
the baseline that employs $\mathrm{A}^*$ combined with suboptimal node selection and branching rules.
As explained in \cref{lcsp:sec:dsp}, this is caused by the way \gls{lpa} handles underconsistent vertices.

\Gls{lpan} handles all underconsistent vertices in a separate preprocessing step, ensuring that only overconsistent vertices are encountered in the main search phase. 
Due to this modification, \gls{lpan} requires 55.38\% fewer arc relaxations than a standard $\mathrm{A}^*$ search.
When considering only the 70 most difficulty instances, the performance gap widens further with \gls{lpan} requiring 63.56\% fewer arc relaxations.

In the \gls{mip} community, it is established that using \emph{\gls{dfs}} for node selection leads to a higher similarity between consecutive subproblems and thus faster subproblem handling. This effect may even compensate for the usually larger tree size \citep{Achterberg2005}.
When using \gls{lpan} in \cref{lcsp:algo:bb}, such an advantage cannot be observed: 
While \emph{\gls{dfs}} ($\expnumber{73.56}{6}$) requires fewer arc relaxations per \gls{sp} search than \emph{best-first search} ($\expnumber{79.54}{6}$), the effect is  negligible and does not translate into a reduction in the total number of arc relaxations.

\paragraph{Comparison against \glsentryshort{mip}}


\begin{figure}
    \centering
    \includegraphics[]{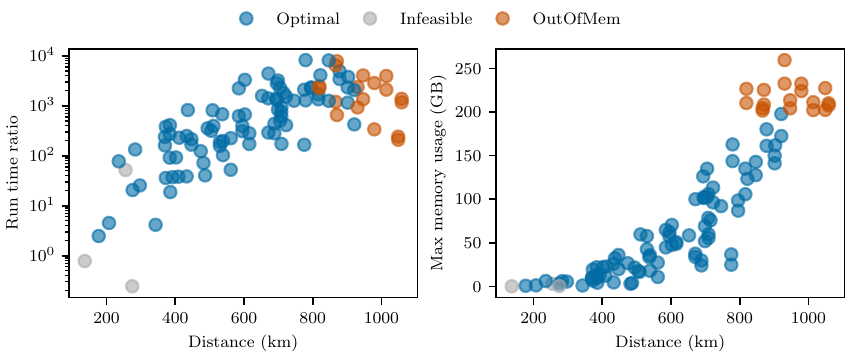}
    \caption[Comparison with MIP approach]{Performance of the \gls{mip} approach on all 106 domestic \glspl{od}.
        Left: Run time of the \gls{mip} approach divided by the run time of \cref{lcsp:algo:bb} using the recommended configuration.
        Right: Maximum memory consumption while solving the \gls{mip}. 
        Colors indicate whether the \gls{mip} approach solved an instance to optimality,
    proved infeasibility or terminated with an \gls{oom} error.}\label{lcsp:fig:mip}
\end{figure}

In the last experiment, we compare the \gls{mip} formulation \tfrmip{} against \cref{lcsp:algo:bb} on the set of domestic \glspl{od}.
For \cref{lcsp:algo:bb}, we use the \emph{recommended configuration} identified in the first three 
experiments, which consists of \emph{best-first search} node selection,
\emph{SUP} branching on graph conflicts and the \gls{lpan} algorithm for solving the \gls{sp} subproblem. 
Aggregated computational results are reported in \cref{lcsp:table:mip}.
In the geometric mean, \cref{lcsp:algo:bb} is more than two orders of magnitude faster than \tfrmip.
Additionally, \cref{lcsp:algo:bb} solved all instances (or proved infeasibility), while \tfrmip{} terminated with
an \gls{oom} error for 12 out of 106 instances.

In \cref{lcsp:fig:mip}, we depict the run time ratio between \tfrmip{} and \cref{lcsp:algo:bb} and the memory consumption of \tfrmip{} for each \gls{od}
over the straight line distance between the origin and destination.
We see that  the run time advantage of \cref{lcsp:algo:bb} grows exponentially with the distance.
Additionally, \tfrmip{} ran out of memory for all \glspl{od} with a distance of over 920 km.

\section{Conclusion}\label{lcsp:sec:conclusion}
We proposed a \gls{bb} algorithm for the \gls{lcsp}.
The algorithm was demonstrated to be suitable for the \gls{fpptfr} on large-scale $3$d instances with numerous mandatory \glspl{tfr}.
The obtained run times outperform a straightforward \gls{mip} formulation by more than two orders of magnitude.

The algorithm allows for the easy adaptation of known techniques from the \gls{mip} and \gls{sat} communities, such as branching rules and unit propagation.
These techniques, however, require careful reevaluation as conventional wisdom no longer applies in the \gls{lcsp} setting.
The strongest branching rule for the \gls{lcsp}, \glsentrylong{sup}, for instance, is generally considered inadequate in the \gls{sat} community.
Carefully choosing an appropriate configuration of branching rule, node selection rule, conflict, and shortest path algorithm yielded
a order-of-magnitude reduction in the number of required arc relaxations compared to a naive baseline configuration.
When considering only the most difficult instances, this reduction improves to 98.06\%.
Here, the best configuration of the \gls{bb} algorithm only requires a run time of \num{129}\,s to perform \num{22.52} \gls{sp} searches in the geometric mean,
making it clearly fit for commercial application.


\section*{Acknowledgement}
This work would not have been possible without our industry partner Lufthansa Systems GmbH and in particular without Anton Kaier and Adam Schienle.
We thank them for the data and for the fruitful discussions we had.
We also thank Marco Blanco for sharing his flight planning tool with us.

\bibliographystyle{elsarticle-num-names}
\bibliography{lit}

\appendix
\section{Computational Results}\label{lcsp:appendix:results}

\begin{table}[H]
\centering
\caption[Computational results for all node selection rules]{Results for all node selection rules.
    We report the geometric mean and maximum of the number of \gls{bb} nodes and \gls{sp} queries  as well as the time spent in total and on \gls{sp} queries.
    In all cases apart from the baseline, \emph{\gls{sup}} branching on graph conflicts and the \gls{lpan} algorithm were used.
    }
\small
\setlength{\tabcolsep}{3pt}
\label{lcsp:table:nodeselection}
\begin{tabular}{lrrrrrrrrr}
\toprule
& \multicolumn{2}{c}{Nodes} & \multicolumn{2}{c}{SP} & \multicolumn{3}{c}{Time} &  \multicolumn{2}{c}{Arc Rel. ($\times \expnumber{1}{6}$)}\\
                            \cmidrule(lr){2-3} \cmidrule(lr){4-5} \cmidrule(lr){6-8} \cmidrule(lr){9-10}
Node Sel. Rule               & Mean & Max & Mean & Max  & Total (s) & SP (s) & SP(\%) & Total & per SP\\
\midrule
DFS & 35.73 & 463 & 26.98 & 371 & 197.40 & 185.51 & 93.98 & 1984.78 & \bf 73.56\\
MostFeasible &  27.92 & 333 & 21.32 & 255 & 178.14 & 169.10 & 94.93 & 1782.94 & 83.65\\
BestFirst &\bf 19.70 &\bf 261 &\bf 15.15 & \bf 201 & \bf 119.66 & \bf 113.67 & 94.99 & \bf 1204.94 & 79.54\\
BestFirstPlunge & 28.85 & 291 & 21.18 & 225 & 164.25 & 155.07 & 94.41 & 1764.73 & 83.33\\
Hybrid &23.06 & 262 & 17.63 & 202 & 140.64 & 133.56 & 94.97 & 1398.18 & 79.29\\
Projection &24.87 & 323 & 19.02 & 250 & 156.64 & 148.48 & 94.79 & 1534.77 & 80.68\\
\midrule
Baseline  & 62.93 & 791 & 60.60 & 780 & 1015.51 & 989.49 & 97.44 & 13804.44 & 227.80\\
\bottomrule
\end{tabular}
\end{table}
\begin{table}[H]
    \centering
    \caption[Computational results for branching rules]{\small Results for all branching rules.
        We report the geometric mean and maximum of the number of \gls{bb} nodes and \gls{sp} queries, the time spent in total and on \gls{sp} queries,
        and the number of arc relaxations.
        In all cases apart from the {baseline}, \emph{best-first search} node selection and the {\gls{lpan}} algorithm were used.
             }
    \label{lcsp:table:branching}
    \small
    \setlength{\tabcolsep}{3pt}
    \begin{tabular}{lrrrrrrrrr}
        \toprule
                                     & \multicolumn{2}{c}{Nodes} & \multicolumn{2}{c}{SP} & \multicolumn{3}{c}{Time} & \multicolumn{2}{c}{Arc Rel. ($\times \expnumber{1}{6}$)}\\
                                      \cmidrule(lr){2-3} \cmidrule(lr){4-5} \cmidrule(lr){6-8} \cmidrule(lr){9-10}
        Branch. Rule               & Mean & Max & Mean & Max  & Total (s) & SP (s) & SP(\%) & Total & Per SP\\
\midrule
SUP & 22.71 & 499 & 21.68 & 308 & 131.85 & 125.47 & 95.16 & 1274.98 & 58.81\\
SUP (Gr.) & 19.70 & 261 & \bf 15.15 & 201 & \bf 120.16 & \bf 114.06 & 94.92 & \bf 1204.94 & 79.54\\
DUP & 21.91 & 499 & 20.72 & 304 & 175.07 & 139.26 & 79.55 & 1343.78 & 64.85\\
DUP (Gr.) & 25.16 & 449 & 19.37 & 358 & 194.12 & 148.21 & 76.35 & 1495.85 & 77.24\\
MOMS & 23.20 & 497 & 22.13 & 412 & 157.76 & 150.79 & 95.58 & 1531.60 & 69.21\\
MOMS (Gr.) & 20.77 & 233 & 17.11 & \bf 177 & 139.20 & 132.78 & 95.39 & 1376.45 & 80.44\\
CVDS & 23.38 & 499 & 22.33 & 362 & 151.10 & 144.27 & 95.48 & 1438.33 & 64.40\\
CVDS (Gr.) & 22.09 & 665 & 19.81 & 552 & 154.95 & 148.52 & 95.85 & 1493.75 & 75.41\\
Clause & 27.57 & 663 & 26.61 & 537 & 158.69 & 150.88 & 95.08 & 1451.69 & 54.56\\
Strong (Gr.) & \bf 15.18 & \bf 129 & 49.28 & 557 & 305.75 & 286.53 & 93.71 & 2481.50 & \bf 50.36\\
\midrule
Baseline  & 62.93 & 791 & 60.60 & 780 & 1015.51 & 989.49 & 97.44 & 13804.44 & 227.80\\
\bottomrule
\end{tabular}
\end{table}
\begin{table}[H]
    \centering
    \caption[Computational results for difficult instances]{Results for all branching rules on the 70 difficult instances, requiring
         more than 300 \gls{bb} nodes for at least one configuration, including the baseline.  
         Reported are the geometric mean and maximum of the number of \gls{bb} nodes and \gls{sp} queries, the time spent in total and on \gls{sp} queries,
         and the number of arc relaxations. 
         In all cases apart from the {baseline}, \emph{best-first search} node selection and the \gls{lpan} algorithm were used.}
    \label{lcsp:table:hard}
    \small
    \setlength{\tabcolsep}{3pt}
    \begin{tabular}{lrrrrrrrrr}
    \toprule
                                      & \multicolumn{2}{c}{Nodes} & \multicolumn{2}{c}{SP} & \multicolumn{3}{c}{Time} & \multicolumn{2}{c}{Arc Rel. ($\times \expnumber{1}{6}$)} \\
                                      \cmidrule(lr){2-3} \cmidrule(lr){4-5} \cmidrule(lr){6-8} \cmidrule(lr){9-10}
    Branch. Rule               & Mean & Max & Mean & Max  & Total (s) & SP (s) & SP(\%) & Total & Per SP \\
        \midrule
SUP & 
45.23 & 499 & 41.86 & 308 & 151.43 & 138.24 & 91.29 & 1230.78 & 29.40\\
SUP (Gr.) &
30.83 & 261 & \bf 22.52 & 201 & \bf 137.92 & \bf 129.29 & 93.75 & 1344.62 & 59.71\\
DUP &
43.05 & 499 & 39.50 & 304 & 261.05 & 147.24 & 56.40 & \bf 1193.25 & 30.21\\
DUP (Gr.) & 
42.64 & 449 & 31.00 & 358 & 242.72 & 172.80 & 71.20 & 1499.56 & 48.37\\
MOMS &
50.08 & 497 & 45.26 & 412 & 192.45 & 175.68 & 91.29 & 1522.01 & 33.63\\
MOMS (Gr.) &
33.88 & 233 & 27.26 & \bf 174 & 161.93 & 152.18 & 93.97 & 1485.67 & 54.49\\
CVDS &
57.49 & 499 & 52.79 & 362 & 182.76 & 162.97 & 89.18 & 1325.06 & 25.10\\
CVDS (Gr.) & 
41.03 & 665 & 35.35 & 552 & 186.07 & 175.72 & 94.44 & 1466.35 & 41.48\\
Clause & 
85.80 & 663 & 79.56 & 537 & 267.12 & 238.16 & 89.16 & 1599.08 & \bf 20.10\\
Strong (Gr.) & 
\bf 19.81 & \bf129 & 58.71 & 557 & 308.93 & 286.44 & 92.72 & 2051.72 & 34.95\\
\midrule
Baseline  & 349.29 & 791 & 323.38 & 780 & 4673.88 & 4535.69 & 97.04 & 62321.27 & 192.72\\
\bottomrule
\end{tabular}
\end{table}
\begin{table}[H]
    \centering
    \caption[Comparison of shortest path algorithms]{Comparison of shortest path algorithms. \emph{Medium instances} refers
        to all instances that require more than 50 \gls{bb} nodes in any configuration, including the baseline.
        We exclude a single \gls{od}, Brindisi to Ontario, that could not be solved by \gls{lpa} within a time limit of five days.
        \emph{Hard instances} refers to all instances that require more than 300 \gls{bb} nodes.
        Reported is the time spent in total and on \gls{sp} queries,
        as well as the number of arc relaxations in total and per \gls{sp} query.
        All experiments were conducted using the \emph{\gls{sup}} branching rule.
        In all cases, apart from \emph{\gls{lpan} (\gls{dfs})}, \emph{best-first-search} node selection was used.
    }
    \label{lcsp:table:dynamicsp}
    \small
    \setlength{\tabcolsep}{3pt}
    \begin{tabular}{lrrrrrrrr}
    \toprule
                          &\multicolumn{4}{c}{Medium Instances} & \multicolumn{4}{c}{Hard Instances} \\
                               \cmidrule(lr){2-5} \cmidrule(lr){6-9}
                           & \multicolumn{2}{c}{Time (s)}  & \multicolumn{2}{c}{Arc Rel. ($\times \expnumber{1}{6}$)} & 
                           \multicolumn{2}{c}{Time (s)}  & \multicolumn{2}{c}{Arc Rel. ($\times \expnumber{1}{6}$)}\\
                               \cmidrule(lr){2-3} \cmidrule(lr){4-5} \cmidrule(lr){6-7} \cmidrule(lr){8-9}
                          & Total     & SP     & Total       & Per SP & Total          &  SP & Total    &   Per SP\\
        \midrule                                                                                               
        $\mathrm{A*}$        & 212.39 & 204.82   & 2700.48 & 178.26  &   297.37 & 285.19 & 3689.52  & 163.83\\
        \gls{lpa}            &1179.49 & 1168.82 & 15936.05 & 1052.94 &   1147.00 & 1134.41& 15557.53 & 690.82\\
        \gls{lpan}           & \bf 127.52 & \bf 120.91   & \bf 1204.91 & 79.54   &   \bf 146.95 & \bf 137.80 & \bf 1344.62 & \bf 59.71\\   
        \gls{lpan} (DFS)     & 197.40 & 185.51   & 1984.78 & \bf 73.56   &   234.78 & 217.47 & 2327.60 & 62.13\\            
        \midrule
        Baseline             &1014.28 & 988.28   & 13787.98 & 227.63 & 4673.88 & 4535.69 &  62321.27 & 192.72\\
    \bottomrule
    \end{tabular}
\end{table}

\begin{table}[H]
    \caption[Comparison between the \glsentryshort{mip} and the hybrid approach]{
        Performance of the \gls{mip} approach compared against \cref{lcsp:algo:bb} using both the baseline configuration 
        and the recommended configuration.
        Reported are the primal-dual gap, the run time in seconds and the number of instances solved to optimality, proven infeasible
        or terminated due to a \gls{oom} error.
        For the \gls{mip} formulation \tfrmip, we also report the geometric mean of the number of variables and constraints.}
    \label{lcsp:table:mip}
    \centering
    \small
    \setlength{\tabcolsep}{3pt}
    \begin{tabular}{lrrrrrrrr}
        \toprule
                    & Gap (\%)  & Time (s)    & Opt &Infeas& \gls{oom} & Vars ($\times\expnumber{1}{6}$)& Cons ($\times\expnumber{1}{6}$)\\
        \midrule
        \tfrmip{}                          & 0.01 & 2737.82 & 85  & 3   & 18 & 17.48 & 20.53 \\
        Alg. \ref{lcsp:algo:bb} (Base.)    & 0.0  & 20.34   & 103 & 3   & 0  & --- & ---     \\
        Alg. \ref{lcsp:algo:bb} (Rec.)     & 0.0  & 6.39    & 103 & 3   & 0  & --- & ---     \\
        \bottomrule
    \end{tabular}
\end{table}

\end{document}